\documentclass[journal]{IEEEtran}
\ifCLASSINFOpdf
   \usepackage[pdftex]{graphicx}
\else
\fi

\usepackage{url}
\usepackage{bm}
\usepackage{multirow}
\usepackage{tabularx}
\usepackage{color}
\usepackage{hyperref}
\usepackage{caption}
\usepackage{soul}

\usepackage{booktabs}

\begin{document}
%
\title{Video Salient Object Detection \\Using Spatiotemporal Deep Features}
%
%
%

\author{Trung-Nghia Le and Akihiro Sugimoto
\thanks{Trung-Nghia Le is with the Department of Informatics, SOKENDAI (Graduate University for Advanced Studies), Tokyo, Japan. {\it e-mail: ltnghia@nii.ac.jp}}
\thanks{Akihiro Sugimoto is with the National Institute of Informatics, Tokyo, Japan. {\it e-mail: sugimoto@nii.ac.jp}}}

\markboth{IEEE TRANSACTIONS ON IMAGE PROCESSING}%
{Video Salient Object Detection using Spatiotemporal Deep Features}
%



\maketitle

\begin{abstract}

This paper presents a method for detecting salient objects in videos where temporal information in addition to spatial information is fully taken into account. Following recent reports on the advantage of deep features over conventional hand-crafted features, we propose a new set of SpatioTemporal Deep (STD) features that utilize local and global contexts over frames. We also propose new SpatioTemporal Conditional Random Field (STCRF) to compute saliency from STD features. STCRF is our extension of CRF to the temporal domain and describes the relationships among neighboring regions both in a frame and over frames. STCRF leads to temporally consistent saliency maps over frames, contributing to accurate detection of salient objects' boundaries and noise reduction during detection. Our proposed method first segments an input video into multiple scales and then computes a saliency map at each scale level using STD features with STCRF. The final saliency map is computed by fusing saliency maps at different scale levels. Our experiments, using publicly available benchmark datasets, confirm that the proposed method significantly outperforms state-of-the-art methods. We also applied our saliency computation to the video object segmentation task, showing that our method outperforms existing video object segmentation methods.

\end{abstract}


%
\IEEEpeerreviewmaketitle

\section{Introduction} \label{section:introduction}

Salient object detection from videos plays an important role as a pre-processing step in many computer vision applications such as video re-targeting~\cite{Taoran-ICIP2010}, object detection~\cite{Guo-Neurocomputing2014}, person re-identiﬁcation~\cite{Zhao-CVPR2013}, and visual tracking~\cite{Stalder-ACCV2012}. Conventional methods for salient object detection often segment each frame into regions and artificially combine low-level (bottom-up) features (e.g., intensity~\cite{Zhu-CVPR2014}, color~\cite{Zhu-CVPR2014}, edge orientation~\cite{Nghia-PSIVT2015}) with heuristic (top-down) priors (e.g., center prior~\cite{Zhou-CVPR2014}, boundary prior~\cite{Zhu-CVPR2014}, objectness~\cite{Nghia-PSIVT2015}) detected from the regions. Low-level features and priors used in the conventional methods are hand-crafted and are not sufficiently robust for challenging cases, especially when the salient object is presented in a low-contrast and cluttered background. Although machine learning based methods have been recently developed~\cite{Long-CVPR2013}\cite{Jiang-CVPR2013}\cite{Liu-PAMI2011}, they are primary for integrating different hand-crafted features~\cite{Jiang-CVPR2013}\cite{PJiang-ICCV2013} or fusing multiple saliency maps generated from various methods~\cite{Long-CVPR2013}. Accordingly, they usually fail to preserve object details when the salient object intersects with the image boundary or has similar appearance with the background where hand-crafted features are often unstable.

Recent advances in deep learning using Deep Neural Network (DNN) enable us to extract visual features, called deep features, directly from raw images/videos. They are more powerful for discrimination and, furthermore, more robust than hand-crafted features~\cite{Taylor-ECCV2010}\cite{Girshick-CVPR2014}\cite{Tran-ICCV2015}. Indeed, saliency models for videos using deep features~\cite{Li-CVPR2016}\cite{Liu-CVPR2016}\cite{Lee-CVPR2016} have demonstrated superior results over existing works utilizing only hand-crafted features. However, they extract deep features from each frame independently and employ frame-by-frame processing to compute saliency maps, leading to inaccuracy for dynamically moving objects. This is because temporal information over frames is not taken into account in computing either deep features or saliency maps. Incorporating temporal information in such computations should lead to better performance. 

\begin{figure}[t]
    \includegraphics[width=1\linewidth]{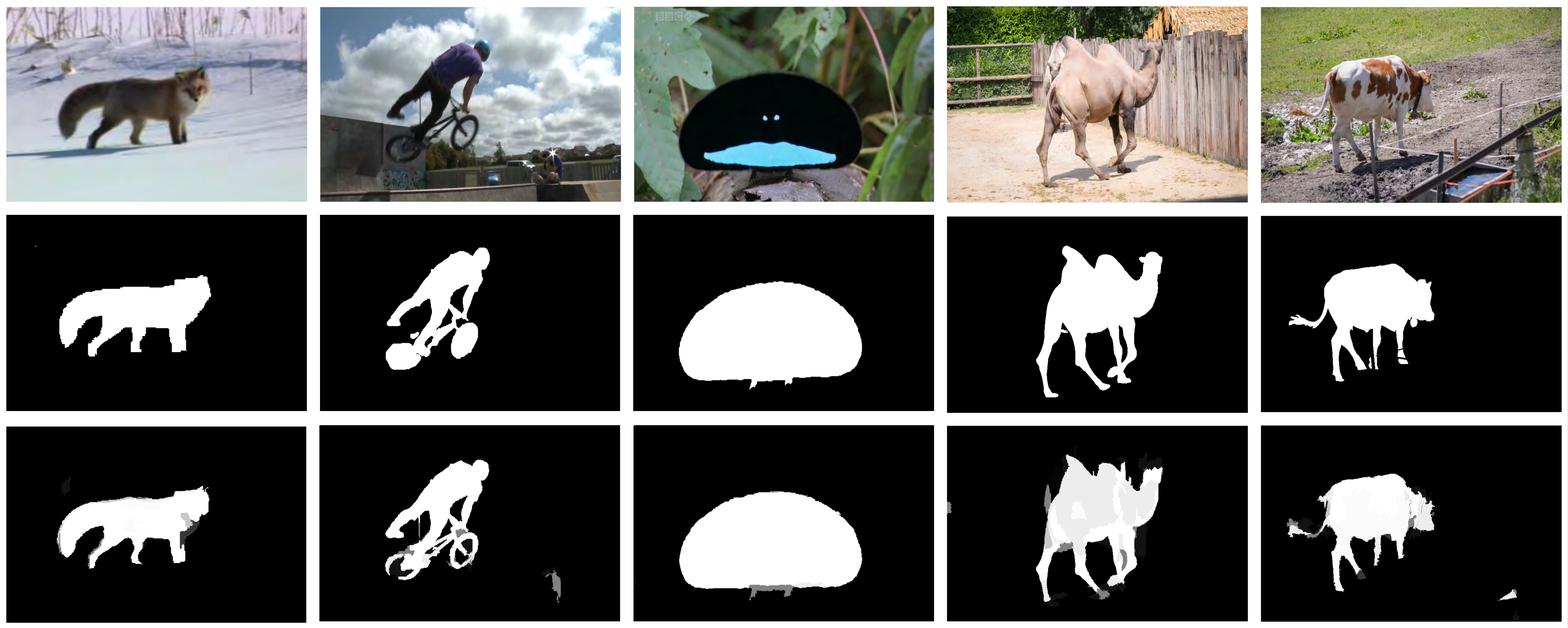}
    \caption{Examples of results obtained by our proposed method. Top row images are original video frames, followed by the ground truth and corresponding saliency maps obtained using our method.}
    \label{img:examples}
\end{figure}

Computed saliency maps do not always accurately reflect the shapes of salient objects in videos. To segment salient objects as accurately as possible while reducing noise, dense Conditional Random Field (CRF)~\cite{Li-CVPR2016}\cite{Hou-CVPR2017}, a powerful graphical model to globally capture the contextual information, has been applied to the computed saliency maps, which results in improvement in spatial coherence and contour localization. However, dense CRF is applied to each frame of a video separately, meaning that only spatial contextual information is considered. Again, temporal information over frames should be taken into account for better performance.

Motivated by the above observation, we propose a novel framework using spatiotemporal information as fully as possible for salient object detection in videos. We introduce a new set of SpatioTemporal Deep (STD) features that utilize both local and global contexts over frames. Our STD features consist of local and global features. The local feature is computed by aggregating over frames deep features, which are extracted from each frame using a region-based Convolutional Neural Network (CNN)~\cite{Girshick-CVPR2014}. The global feature is computed from a temporal-segment of a video using a block-based\footnote{In contrast to  the region-based CNN working on spatial segments in each frame, the block-based CNN works on a sequence of frames of a video.} CNN~\cite{Tran-ICCV2015}.

We also introduce the SpatioTemporal CRF (STCRF), in which the spatial relationship between regions in a frame as well as temporal consistency of regions over frames is formally described using STD features. Our proposed method first segments an input video into multi-scale levels, and then at each scale level, extracts STD features and computes a saliency map. The method then fuses saliency maps at different scale levels into the final saliency map. 
Extensive experiments on public benchmark datasets for video saliency confirm that our proposed method significantly outperforms the state-of-the-arts. Examples of saliency maps obtained by our method are shown in Fig.\ref{img:examples}. We also apply our method to video object segmentation and observe that our method outperforms existing methods. 

The rest of this paper is organized as follows. We briefly review and analyze related work in Section \ref{section:related_work}. Then, we present in detail our proposed method in Section \ref{section:proposed_method}. Our experiments are discussed in Sections \ref{section:setting} and \ref{section:experiments}. In Section \ref{section:application}, we present an application of our proposed method to video object segmentation. Section \ref{section:conclusion} presents conclusion and future work. 
We remark that this paper extends the work reported in~\cite{Nghia-DeLIMMA2017}. Our extensions in this paper are building a new STCRF model utilizing CNN instead of Random Forest (Section \ref{section:STCRF_model}), adding more experiments (Section \ref{section:experiments}), and an application of our method to video object segmentation (Section \ref{section:application}).

\section{Related Work} \label{section:related_work}

Here we briefly survey features used for salient object detection in videos, and saliency computation methods.

\subsection{Features for Salient Object Detection}

Saliency computation methods for videos using hand-crafted features are mostly developed from traditional saliency models for still images by incorporating motion features to deal with moving objects~\cite{Nghia-PSIVT2015}\cite{Zhou-CVPR2014}\cite{Liu-PAMI2011}\cite{Rahtu-ECCV2010}. 
Motion features commonly used include optical flow~\cite{Nghia-PSIVT2015}\cite{Zhou-CVPR2014}\cite{Rahtu-ECCV2010}, trajectories of local features~\cite{Liu-PAMI2011}\cite{Zhai-MM2006}, gradient flow field~\cite{Wang-TIP2015}, and temporal motion boundary~\cite{Wang-CVPR2015}; they are utilized to detect salient objects in videos. Xue et al.~\cite{Xue-ICASSP2012}, on the other hand, sliced a video along $X$--$T$ and $Y$--$T$ planes to separate foreground moving objects from backgrounds. However, hand-crafted features have limitation in capturing the semantic concept of objects. Accordingly, these methods often fail when the salient object crosses the image boundary or has similar appearance with the background.

Several existing methods~\cite{Li-CVPR2016}\cite{Li-CVPR2015} for saliency computation using deep features, on the other hand, utilize superpixel segmentation to extract region-level deep features in different ways (e.g., feeding regions into a CNN individually to compute deep features~\cite{Li-CVPR2015} or pooling a pixel-level feature map into regions to obtain region-level deep features~\cite{Li-CVPR2016}).
To exploit the context of a region in multiple scales, multi-scale deep features of the region are extracted by changing the window size~\cite{Li-CVPR2015}. Li et al.~\cite{Li-CVPR2015} fused multi-scale deep features of a region of interest to compute the saliency score for the region using a two-layer DNN. Lee et al.~\cite{Lee-CVPR2016} integrated 
hand-crafted features into deep features to improve accuracy for salient object detection. More precisely, they concatenated an encoded low-level distance map and a high-level feature map from CNN to enrich information included in the extracted feature map. The region-level feature map and the pixel-level feature map are also integrated into the saliency model to enhance accuracy of detected object boundaries~\cite{Li-CVPR2016}. In end-to-end deep saliency models~\cite{Liu-CVPR2016}\cite{Wang-ECCV2016}, pixel-based deep features are enhanced by their context information through recurrent CNNs. 

Saliency models using deep features have demonstrated state-of-the-art performance in salient object detection and significantly outperformed existing works utilizing only hand-crafted features. However, in almost all existing saliency models, temporal information over frames is not taken into account in deep features, leading to inaccuracy for dynamically moving objects.
Though Wang et al.~\cite{Wang-TIP2018} very recently proposed a fully convolutional network (FCN) having a pair of frames as its input for video saliency computation, a pair of frames is too short to exploit the temporal domain.
Therefore, effectively mining correlation inherent in the spatial and temporal domains into powerful deep features for saliency computation is still an open problem.

\begin{figure*}[t]
    \begin{center}
        \includegraphics[width=0.9\linewidth]{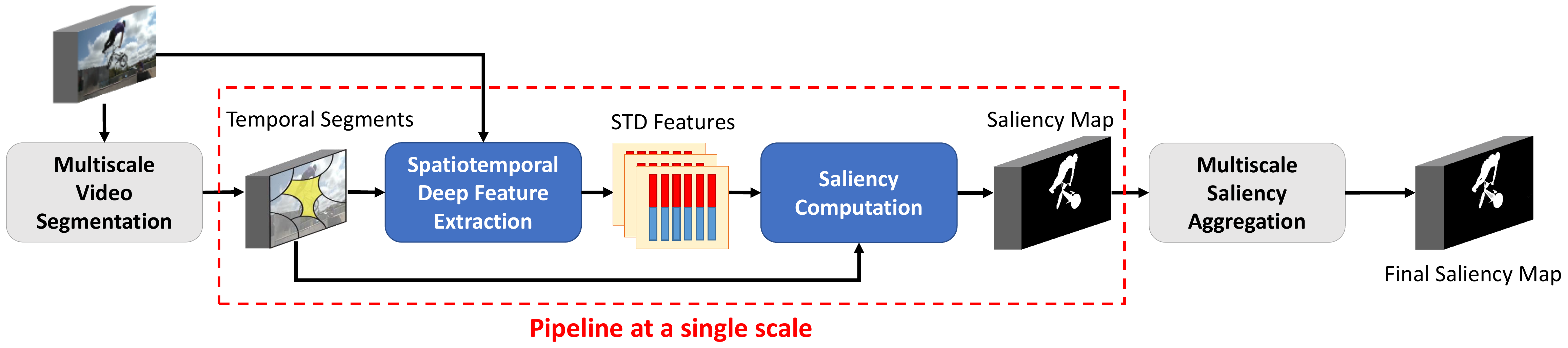}
        \end{center}
        \centering
    \caption{Pipeline of the proposed method (brighter means more salient in the final saliency map).}
    \label{fig:overview}
\end{figure*}

\subsection{Saliency Computation Methods}

The salient object detection approach using deep models ~\cite{Li-CVPR2016}\cite{Liu-CVPR2016}\cite{Hou-CVPR2017}\cite{Wang-ECCV2016}\cite{Xi-TIP2015} computes saliency scores directly from FCNs. In these deep models, recurrent layers~\cite{Liu-CVPR2016}\cite{Wang-ECCV2016} and skip connections~\cite{Liu-CVPR2016}\cite{Hou-CVPR2017} are utilized to enhance the contextual information of deep feature maps to improve the accuracy of saliency computation. However, these methods focus on frame-by-frame processing without considering any temporal information in videos. 
In addition, they still do not detect boundaries of salient objects accurately. A refinement post-processing step is usually required to improve accuracy of detected object boundaries.

Spatial CRF has the capability to relate local regions in order to capture global context, and has been commonly used for refinement in semantic segmentation~\cite{Shimoda-ECCV2016} and for saliency computation~\cite{Li-CVPR2016}\cite{Hou-CVPR2017}. Dense CRF~\cite{Philipp-NIPS2011} is used as a post-processing step to refine the label map generated from CNN to improve the performance of semantic segmentation~\cite{Shimoda-ECCV2016}. Shimoda et al.~\cite{Shimoda-ECCV2016} developed a weakly supervised semantic segmentation method using a dense CRF to refine results from distinct class saliency maps. The dense CRF is incorporated into the saliency map computed from the CNN to improve spatial coherence and contour localization~\cite{Li-CVPR2016}\cite{Hou-CVPR2017}. Though spatial information is successfully utilized using CRFs in these methods, how to deal with temporal information is left unanswered, which is crucial for videos.

Dynamic CRF (DCRF)~\cite{Yang-CVPR2005} is an extension of the spatial CRF toward to the spatiotemporal domain to exploit both spatial and temporal information in videos. DCRF is constructed from consecutive video frames, where each pixel connects to its neighboring pixels in both space (i.e., the same frame) and time (i.e., the next frame and the previous frame). DCRF has been used to enhance both spatial accuracy and temporal coherence for object segmentation~\cite{Yang-CVPR2005}\cite{Yang-PAMI2006}\cite{Yi-CVPR2016} and saliency computation~\cite{Liu-PAMI2011} in videos. Yi et al.~\cite{Yi-CVPR2016} proposed a framework using DCRF to improve fence segmentation in videos. Wang et al.~\cite{Yang-CVPR2005}\cite{Yang-PAMI2006} applied DCRF to object segmentation and moving shadow segmentation in indoor scenes in videos. SIFT flow features were incorporated into DCRF to detect salient objects from videos~\cite{Liu-PAMI2011}. However, DCRF is a pixel-level dense graph; thus it is usually constructed using only two successive frames due to large memory consumption. In addition, since the energy function of DCRF is defined using the combination of classical hand-crafted features such as color and optical flow, DCRF is not capable of exploiting spatial and temporal information semantically. Our proposed STCRF differs from DCRF in that STCRF is defined over regions using STD features only, so that it is capable of dealing with more successive frames and exploiting spatial and temporal information semantically with less computational cost.

Different from these existing methods, our proposed method utilizes spatiotemporal information as much as possible when both extracting deep features and computing saliency maps. More precisely, our method uses STD features computed from the spatiotemporal domain together with STCRF constructed in the spatiotemporal domain to produce accurate saliency maps. Our method thus accurately detects boundaries of salient objects by removing irrelevant small regions.

\section{Proposed Method} \label{section:proposed_method}

\subsection{Overview}

Our goal is to compute a saliency map to accurately segment salient objects in every frame from an input video while fully utilizing information along the temporal dimension. Figure \ref{fig:overview} illustrates the pipeline of our proposed method. 

We segment an input video at multiple scale levels and compute a saliency map at each scale level at each frame, and then aggregate all saliency maps at different scale levels at each frame into a final saliency map. This follows our intuition that objects in a video contain various salient scale patterns and an object at a coarser scale level may be composed of multiple parts at a finer scale level.

In this work, we employ the video segmentation method~\cite{Liu-CVPR2011} at multiple scale levels. We first specify the number of initial superpixels to define a scale level. For each scale level, we then segment each frame into initial superpixels using entropy-rate superpixel segmentation~\cite{Liu-CVPR2011}. Similar superpixels at consecutive frames are then grouped and connected across frames to have temporal segments using parametric graph partitioning~\cite{yu-ICCV2015}. 
By specifying different numbers of initial superpixels, we obtain multiple scale temporal segments (we set four numbers to have four scale levels in our experiments as discussed later). We remark that each scale level has a different number of segments, which are defined as (non-overlapping) regions.

The final saliency map is computed by taking the average value of saliency maps over different scale levels. In the following subsections, we explain how to compute a saliency map at a scale level. We remark that a saliency map in this section indicates the saliency map at a scale level unless explicitly stated with "final."

\subsection{Spatiotemporal Deep Feature Extraction}

\begin{figure*}[t]
    \begin{center}
        \includegraphics[width=0.9\linewidth]{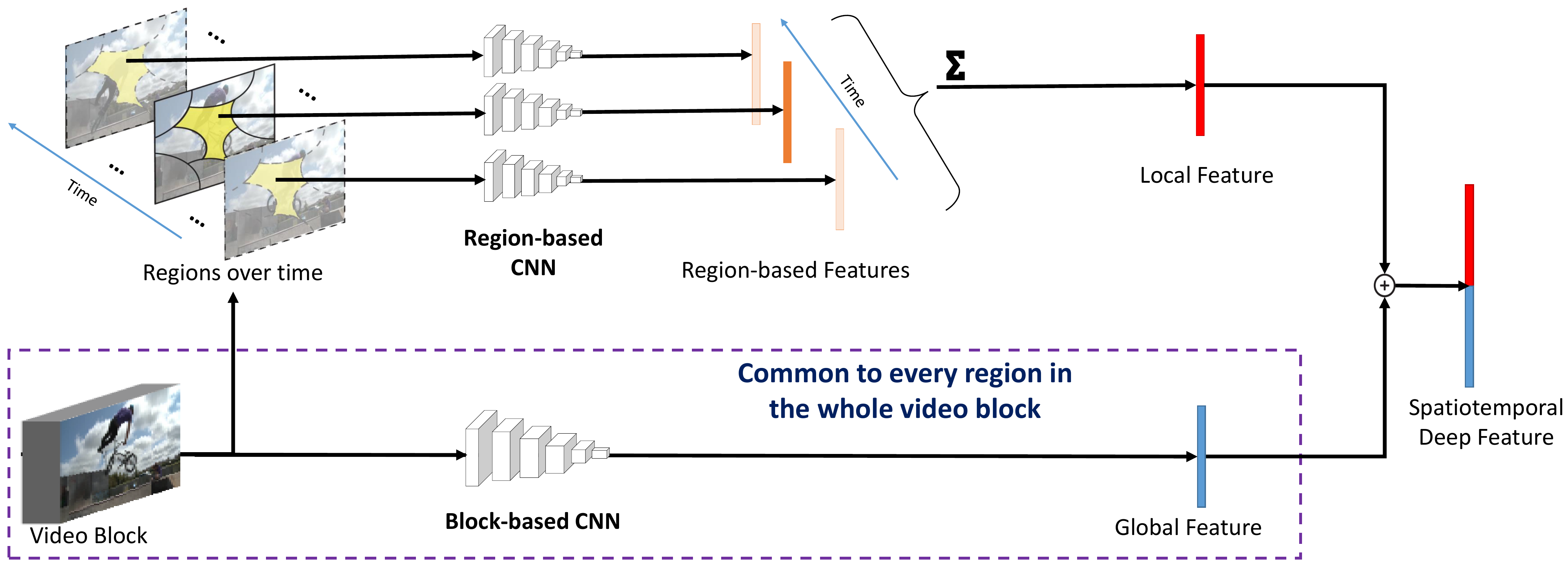}
        \end{center}
        \centering
    \caption{Spatiotemporal deep (STD) feature extraction for a region. 
    A region (yellow) in a frame of a video block is fed to the region-based CNN to have the region-based feature of the (yellow) region in the frame.
    Region-based features over the frames of the video block are aggregated to have the local feature of the region.  On the other hand, the video block is fed to the block-based CNN to have the global feature.
    The local feature of the region and the global feature are concatenated to form the spatiotemporal deep (STD) feature of the region.}

    \label{fig:feature_extraction}
\end{figure*}

For each region (segment) at each frame, our proposed STD feature is computed by concatenating a local feature and a global feature. The local feature is extracted using a region-based CNN followed by aggregation over frames, while the global feature is computed using a block-based CNN whose input is a sequence of frames of the video. The STD feature extraction for a region is illustrated in Fig.~\ref{fig:feature_extraction}.

\subsubsection{Local Feature Extraction}

A region at each frame, which is defined from a temporal segment at a frame, is fed into a region-based CNN to extract its region-based feature which is with a dimension of 4096. As our region-based CNN, we use the publicly available R-CNN model\footnote{R-CNN runs at the original resolution of its input region while Fast R-CNN~\cite{Girshick-ICCV2015}, Faster R-CNN~\cite{Ren-NIPS2015}, and Mask R-CNN~\cite{Kaiming-ICCV2017} require to reduce the resolution of the region to adapt their architectures. This resolution reduction may eliminate small regions. We thus used R-CNN.}~\cite{Girshick-CVPR2014} that was pre-trained on the ImageNet ILSVRC-2013 challenge dataset~\cite{Russakovsky-IJCV2015}.

The region-based feature contains the local context of the region but does not contain temporal information because it is computed frame-by-frame. In order to incorporate temporal information, for a region, we aggregate its region-based features over a sequence of frames, resulting in the consistent local feature over time for the region. It is important to remark that we use only neighboring frames whenever a region-of-interest is present. Thus, the number of frames used for this aggregation may change depending on the region.

Just uniformly averaging region-based features over frames is not wise because pixels vary over time due to lossy compression, degrading accuracy of corresponding regions across frames. This degradation increases with larger time increments across frames.  We thus linearly combine region-based features at neighboring frames, similarly to~\cite{Nghia-PSIVT2015}, using weights modeled by a Gaussian distribution centered at the frame from which we compute local features. With these weights, region-based features at frames with large temporal distance to a frame of interest will contribute less to the computation of local features of the frame:
the local feature $ F_{\rm L}(i,t)$ of a region $i$ at frame $t$ is extracted by
\begin{eqnarray}
F_{\rm L}(i,t) & = & \frac{1}{\Psi}\sum\limits_{t' = t - {\raise0.5ex\hbox{$\scriptstyle k$}
\kern-0.1em/\kern-0.15em
\lower0.25ex\hbox{$\scriptstyle 2$}}}^{t + {\raise0.5ex\hbox{$\scriptstyle k$}
\kern-0.1em/\kern-0.15em
\lower0.25ex\hbox{$\scriptstyle 2$}}} {{\cal G}\left( {t'|t,{\sigma ^2}} \right)f(i,t')},
\end{eqnarray} 
where 
${\cal G}\left( {t'|t,{\sigma^2}} \right)$ is a Gaussian distribution with mean $t$ and standard deviation $\sigma=2$ expressing distribution of temporal weights,
$f(i,t')$ is the region-based feature of region $i$ at frame $t'$, and 
$\Psi=\sum_{t' = t - k/2}^{t + k/2}{{\cal G}\left( {t'|t,{\sigma ^2}} \right)}$ (normalizing factor). 
$k+1$ is the number of frames where the region $i$ is always present.

In this work, we set $k=16$ by default.  This is because almost all regions at a frame are present in the next (previous) 8 successive frames.  For a region that disappears during the next 7 successive frames or that newly appears during the previous 7 successive frames, we first identify the maximum number of successive frames in which the region is always present in the previous and the next directions and use this number as $k$ for the region. 
For example, if a region in a frame appears from 3 frames before and disappears in 2 frames after, then we set $k=4(=2 \times 2)$ for this region.

\subsubsection{Global Feature Extraction}

To compute a global feature, we feed a video block (sequence of frames) of a video into a block-based CNN. The global feature obtained in this way takes its temporal consistency into account in its nature. As our block-based CNN, we employ the C3D model~\cite{Tran-ICCV2015} pre-trained on the Sports-1M dataset~\cite{Karpathy-CVPR2014}, which is known to be effective for extracting spatiotemporal features for action recognition. As an input video block, frame $t$ is expanded into both directions in the temporal domain to obtain a 16-frame sequence as suggested by~\cite{Tran-ICCV2015}. For each input block, we feed it into the pre-trained C3D model only once and assign the extracted global feature $F_{\rm G}(t)$ with a dimension of 4096 identical to all the regions in the block. This distributes the global context to each region and, at the same time, reduces the computational cost. 

Finally, for a region $i$ of a frame $t$, we concatenate its local and global feature vectors to obtain its STD feature vector $F(i,t)$ whose dimension is $4096 \times 2$: $F(i,t)=F_{\rm L}(i,t) \oplus F_{\rm G}(t)$ (cf. Fig.~\ref{fig:feature_extraction}).

\subsection{Saliency Computation Using SpatioTemporal CRF}

\begin{table}[t]
\centering
\caption{Architecture of the our F-DNN.}
\label{tab:FDNN}
\small
\begin{tabular}{clc}
\toprule
\textbf{No} & \textbf{Layer} & \textbf{Output Channel} \\ \midrule
0 &  STD Feature Input & 8192 \\ \midrule
1 &  Fully Connected & 2048 \\
2 &  ReLU & 2048 \\
3 &  Dropout & 2048 \\ 

4 &  Fully Connected & 2048 \\
5 &  ReLU & 2048 \\
6 &  Dropout & 2048 \\ 

7 &  Fully Connected & 2048 \\
8 &  ReLU & 2048 \\
9 &  Dropout & 2048 \\ 

10 &  Fully Connected & 1024 \\
11 &  ReLU & 1024 \\
12 &  Dropout & 1024 \\ 

13 & Fully Connected & 1024 \\
14 & ReLU & 1024 \\ 
15 & Dropout & 1024 \\ 

16 & Fully Connected  & 1024 \\
17 & ReLU & 1024 \\

18 & Fully Connected & 2 \\
\bottomrule
\end{tabular}
\end{table}

CRF is used to improve accuracy (particularly in object boundaries) of the saliency map while reducing noise because CRF captures the spatial relationship between regions in a frame.  We extend CRF toward the temporal domain to have the ability to capture temporal consistency of regions over frames as well.  
We call our extended CRF, SpatioTemporal CRF (STCRF in short).

\subsubsection{STCRF Graph Construction} \label{section:STCRF_model}

\begin{figure*}[t]
    \centering
        \includegraphics[width=0.9\linewidth]{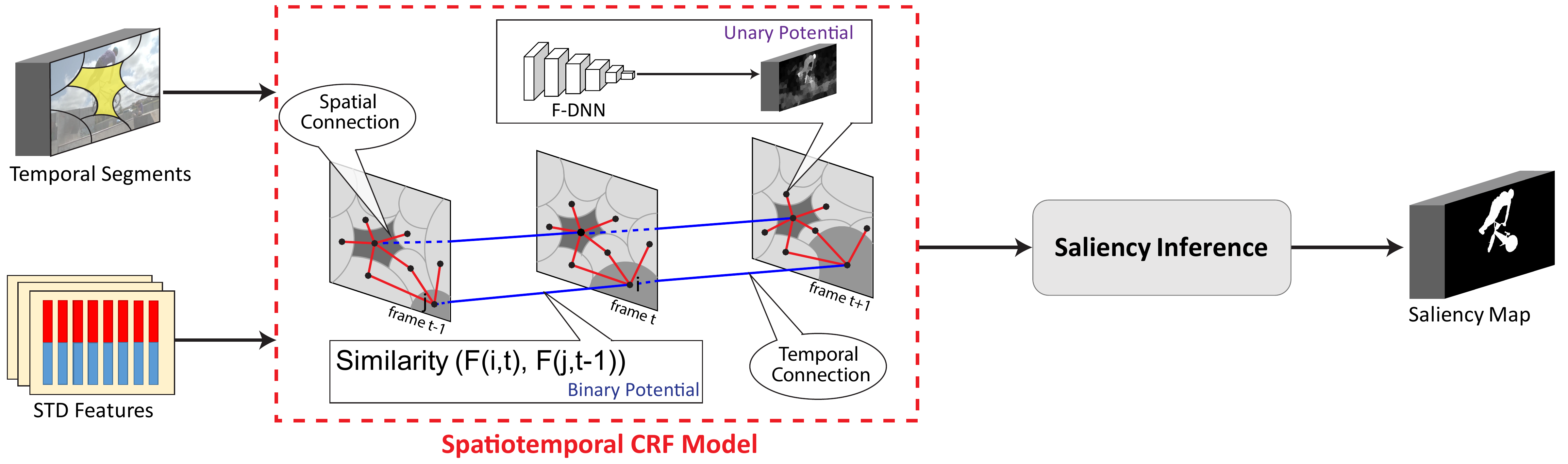}
    \caption{Saliency computation pipeline for a video block based on a graphical model.}
    \label{fig:graphical_model}
\end{figure*}

For temporal segments of a video block, we construct a STCRF graph. Each vertex of the graph represents a region, which is defined from a temporal segment at a frame, in the block. Each edge of the graph, on the other hand, represents the neighboring relationship between regions in space or in time. Considering all the neighboring relationships, however, leads to a dense graph especially when the video volume is large, and the constructed graph becomes practically useless in considering memory consumption and processing time in the inference process. We therefore employ edges that only represent adjacency relationship (cf. Fig.\,\ref{fig:graphical_model}). Furthermore, we partition the video into a sequence of consecutive blocks so that inference in each block is performed separately. 

In the experiments, an input video is decomposed into overlapping blocks with a fixed size where the overlapping rate is 50\%. We note that each block length is equal to 16 frames (see Section \ref{section:ws}). The saliency score of a region is refined by uniformly averaging saliency scores of the region over all the blocks that contain the region. This reduces processing time while keeping accuracy. 

\subsubsection{Energy Function for STCRF} \label{section:energy_function}

We define the energy function of the STCRF so that probabilistic inference is realized by minimizing the function.
The energy function $E$ has a video block (with its temporal segments) $\bm{x}$ as its input.
$E$ is defined by the unary and the binary terms with labels representing foreground/background
$\bm{l}=\{ l_i \in \{0,1\} | i \in {\cal V} \}$
where $l_i$ is the label for region $i$, and ${\cal V}$ is the set of vertices, i.e., regions in ${\bm x}$:
\begin{eqnarray}
E(\bm{l}, \bm{x}; \bm{\theta}) & = &  
\sum_{i \in {\cal V}} {\psi_{\rm u} (l_i;\theta_{\rm u} )} +
\sum_{(i,j) \in {\cal E}} {\psi_{\rm b} (l_i,l_j;\theta_{\rm b} )},
\label{equation:energy}
\end{eqnarray} 
where ${\psi_{\rm u}}$ and ${\psi_{\rm b}}$ are the unary and binary potentials given below. $\cal E$ is the set of edges of the STCRF graph. 
$\bm{\theta} =(\theta_{\rm u}, \theta_{\rm b})$ is the model parameter.

\noindent\textbf{{Unary potential}: }

The unary potential for region $i$ is defined using the label of the region:
\begin{eqnarray}
\psi_{\rm u}\left( l_i ; \theta_{\rm u} \right) & =  & \theta_{\rm u}\, {\omega}(F(i,t_i)),
\end{eqnarray} 
where $t_i$ is the frame in which region $i$ exists, and $\omega$ is a function estimating the probability of the region being the foreground.

To compute $\omega$, i.e., the probability of the region being the foreground, we employ the DNN proposed by Wang et al.~\cite{LWang-CVPR2015} and modify it for our problem (cf. Table \ref{tab:FDNN}). Namely, right before the last fully connected layer of the original network, we add a dropout layer and a fully connected layer followed by a Rectified Linear Unite (ReLU)~\cite{Nair-ICML2010} layer (Nos. 15, 16, and 17 in Table \ref{tab:FDNN}) to increase the depth of the network. We then appropriately change the output channel of the first fully connected layer (Nos. 1, 2, and 3).

Hence, our used network, called Foreground-Deep Neural Network (F-DNN in short), consists of 7 fully connected layers. Each layer executes a linear transformation followed by the ReLU operator. Dropout operations with the ratio of 0.5 are applied after ReLU layers during the training process to avoid overfitting. To the input STD feature having 8192 channels, the numbers of output channels gradually reduce to 2048 at the first three fully connected layers and to 1024 at the next three layers. The last fully connected layer has two output channels representing foreground and background classes.

\noindent\textbf{{Binary potential}: }

The binary potential provides the deep feature based smoothing term that assigns similar labels to regions with similar deep features. Depending on spatial adjacency or temporal adjacency, the potential is differently formulated with further separation of $\theta_{\rm b}$ into $\theta_{\rm bs}$ and $\theta_{\rm bt}$:

\begin{eqnarray}
\psi _{\rm b} \left( {l_i},{l_j};\theta_{\rm b} \right) & = &  
\left\{ 
   \begin{array}{cl}
    \theta_{\rm bs}\,{\Phi}_{\rm bs}\left( {l_i},{l_j}\right) & (i,j) \in {\cal E}_{s} \\
    \theta_{\rm bt}\,{\Phi}_{\rm bt}\left( {l_i},{l_j}\right) & (i,j) \in {\cal E}_{t}
\end{array},
\right.
\end{eqnarray} 
where ${\cal E}_{\rm s}$ and ${\cal E}_{\rm t}$ respectively denote the set of edges representing spatial adjacency and that representing temporal adjacency. Note that ${\cal E}={\cal E}_{\rm s}\cup {\cal E}_{\rm t}$ and ${\cal E}_{\rm s} \cap {\cal E}_{\rm t} = \emptyset$. 

$\Phi_{\rm bs}$ and $\Phi_{\rm bt}$ are spatial smoothness and temporal smoothness between two regions:

\begin{eqnarray}
& & \hspace*{-20pt} \Phi_{\rm bs}(l_i, l_j)  = \nonumber \\ 
& & \hspace*{-15pt}  (1-\delta_{l_i\, l_j})
  D(i,j)^{-1}
  \exp \left( -\beta_{s}{ \left\| F(i,t_i) - F(j,t_j) \right\|^2 } \right), 
  \\
 & & \hspace*{-20pt} {\Phi}_{\rm bt}\left(l_i , l_j\right) =  \nonumber \\
 & & \hspace*{-10pt} \left(1-\delta_{l_i l_j}\right)
 \phi(i,j)
  \exp \left( -\beta_{t}{ \left\| F(i,t_i) - F(j,t_j) \right\|^2 } \right),
\end{eqnarray} 
where 
$\delta$ is the Kronecker delta and
$D(i,j)$ is the Euclidean distance between the two centers of regions $i$ and $j$. 
$\phi$ is the ratio of the area matched by the optical flow inside the two temporally different 
regions~\cite{Papazoglou-ICCV2013}. 
$F(i,t_i)$ is the STD feature of region $i$ (which exists in frame $t_i$). 
The parameters $\beta_s$ and $\beta_t$ are chosen similarly to~\cite{Rother-SIGGRAPH2004} to ensure the exponential term switches appropriately between high and low contrasts:
\begin{eqnarray}
\beta_{s} &  =  & 
\frac{1}{2}( \sum\limits_{(i,j) \in {\cal E}_{s}} {\left\| F(i,t_i) - F(j,t_j) \right\|^2} )^{- 1},  \\
\beta_{t}  & =  &
\frac{1}{2}( \sum\limits_{(i,j) \in {\cal E}_{t}} {\left\| F(i,t_i) - F(j, t_j) \right\|^2})^{ - 1}.
\end{eqnarray} 
We remark that to compute the weight $\phi$, we first count the area transformed from a temporal segment (region) at a frame to its corresponding region at the next frame via optical flow and vice versa, and then take the average of ratios of the areas. 
In the temporal domain, this weight is better than the Euclidean distance because it is independent of the speed of the motion~\cite{Papazoglou-ICCV2013}. In this work, we employ the deep flow method~\cite{Weinzaepfel-ICCV2013} to transfer pixels in the temporal segment.

\subsubsection{Saliency Inference} 

Saliency scores for regions are obtained in terms of labels by minimizing the energy function: 
\begin{eqnarray}
\bm{\hat{l}} & = & \mathop {\arg \min }\limits_{\bm{l}} E(\bm{l},\bm{x}; \bm{\theta}),
\end{eqnarray} 

We minimize $E$ in Eq.~(\ref{equation:energy}) by iterating the Graph Cut method~\cite{Boykov-PAMI2001}, which shows the effectiveness in CRF-based energy minimization~\cite{Cheng-CGF2015}, and is popularly used for object segmentation~\cite{Le-DAVIS2017}\cite{Tsai-CVPR2016}. 
The inputs are initial label $\bm{l}$, block (with its temporal segments) $\bm{x}$, and model parameter $\bm{\theta}$.
The minimization is then executed as the iterative expectation-maximization~\cite{Moon-SPM1996} scheme until convergence. In each iteration, the Graph Cut algorithm~\cite{Boykov-PAMI2001} is used to solve the "Min-Cut/Max-Flow" problem~\cite{Foulds-2012} of the graph, resulting in a new label for each vertex (region). The updated labels are used for the next iteration. After the saliency inference process, we obtain (binary) saliency maps for frames in $\bm{x}$.

\section{Experimental Settings} \label{section:setting}

\subsection{Benchmark Datasets}

We evaluated the performance of our method on three public benchmark datasets: 10-Clips dataset~\cite{Fukuchi-ICME2009}, SegTrack2 dataset~\cite{Li-ICCV2013}, and DAVIS dataset~\cite{Perazzi-CVPR2016}.

\textbf{The 10-Clips dataset}~\cite{Fukuchi-ICME2009} has ten video sequences, each of which contains a single salient object. Each sequence in the dataset has the spatial resolution of $352\times288$ and consists of about 75 frames.

\textbf{The SegTrack2 dataset}~\cite{Li-ICCV2013}
contains 14 video sequences and is originally designed for video object segmentation. A half of the videos in this dataset have multiple salient objects. This dataset is challenging in that it has background-foreground color similarity, fast motion, and complex shape deformation. Sequences in the dataset consist of about 76 frames with various resolutions.

\textbf{The DAVIS dataset}~\cite{Perazzi-CVPR2016} consists of 50 high quality $854 \times 480$ spatial resolution and Full HD 1080p video sequences with about 70 frames per video, each of which has one single salient object or two spatially connected objects either with low contrast or overlapping with image boundary. This is also a challenging dataset because of frequent occurrences of occlusions, motion blur, and appearance changes. In this work, we used only $854 \times 480$ resolution video sequences.

All the datasets contain manually annotated pixel-wise ground-truth for every frame.

\subsection{Evaluation Criteria} \label{section:metrics}

We evaluated the performance using Precision-Recall Curve (PRC), 
F-measure~\cite{Achanta-CVPR2009}, and Mean Absolute Error (MAE). 

The first two evaluation metrics are computed based on the overlapping areas between obtained results and provided ground-truth. Using a fixed threshold between 0 and 255, pairs of $(Precision, Recall)$ scores are computed and then combined to form a PRC. F-measure is a balanced measurement between $Precision$ and $Recall$ as follows: 
\begin{eqnarray}
{F_\beta } & = & \frac{{\left( {1 + {\beta ^2}} \right)Precision \times Recall}}{{{\beta ^2} \times Precision + Recall}}.
\end{eqnarray} 
We remark that we set $\beta^2=0.3$ for F-measure, as suggested by~\cite{Achanta-CVPR2009} so that precision is weighted more heavily. 

For a given threshold, we binarize the saliency map to compute $Precision$ and $Recall$ at each frame in a video and then take the average over frames in the video. After that, the mean of the averages over the videos in a dataset is computed. F-measure is computed from the final precision and recall. When binarizing results for the comparison with the ground truth,  we used F-Adap~\cite{Jia-ICCV2013}, which uses an adaptive threshold $\theta=\mu+\eta$ where $\mu$ and $\eta$ are the mean value and the standard deviation of the saliency scores of the obtained map, and F-Max~\cite{Borji-TIP2015}, which describes the maximum of F-measure scores for different thresholds from 0 to 255.
 
MAE, on the other hand, is the average over the frame of pixel-wise absolute differences between the ground truth $GT$ and obtained saliency map $SM$:
\begin{eqnarray}
{\rm MAE}  & = & \hspace*{-5pt}
\frac{1}{{W \cdot H}}\sum\limits_{x = 1}^W {\sum\limits_{y = 1}^H {\left\| {SM\left( {x,y} \right) - GT\left( {x,y} \right)} \right\|} },
\end{eqnarray}  
where $W$ and $H$ are the width and the height of the video frame. We note that MAE is also computed from the mean average value of the dataset in the same way as F-measure.

\subsection{Implementation Details}

We implemented region-based CNN, block-based CNN, and F-DNN in C/C++ using Caffe~\cite{Jia-MM2014}, and we implemented the other parts in Matlab.  All experiments were conducted on a PC with a Core i7 3.6GHz processor, 32GB of RAM, and GTX 1080 GPU.
We remark that the region-based CNN and the block-based CNN were used without any fine-tuning.

To segment a video, we follow \cite{yu-ICCV2015} as described above. We set the number of initial superpixels at each frame as $\{100, 200, 300, 400\}$ to have four scale levels. The other required parameters are set similarly to \cite{yu-ICCV2015}.
For parameters in STCRF, we empirically set $\theta=(\theta_u, \theta_{bs}, \theta_{bt})=(50, 0.05, 1000)$. All these parameters are fixed throughout experiments.

\subsection{Training F-DNN for Foreground Probability Prediction}

\begin{table}[t]
\centering
\caption{Number of videos used in our experiments.}
\label{tab:dataset}
\begin{tabular}{l|cccc}
\toprule
 \textbf{Dataset}              & \textbf{10-Clips}~\cite{Fukuchi-ICME2009} & \textbf{SegTrack2}~\cite{Li-ICCV2013} & \textbf{DAVIS}~\cite{Perazzi-CVPR2016}& \textbf{Total} \\ \midrule
\textbf{Training} & 6                 & 8                  & 30                  & 44 \\ \midrule
\textbf{Testing}  & 4                 & 6                  & 20                  & 30 \\ \bottomrule
\end{tabular}
\end{table}

In training our F-DNN (see Section \ref{section:energy_function}), we took an approach where we use all three datasets together rather than training our F-DNN for each dataset. This is because each dataset is too small to train a reliable model. Our approach also enables the trained model not to over-fit to a specific dataset.

From each video dataset except for the DAVIS dataset, we chose randomly 60\% (in number) of videos and mixed them into a larger dataset for training while the remaining videos were used for testing each dataset (cf. Table \ref{tab:dataset}). For the DAVIS dataset, we used the training set and the testing set as in the DAVIS Benchmark~\cite{Perazzi-CVPR2016}\footnote{\href{http://davischallenge.org/browse.html}{http://davischallenge.org/browse.html}}. We thus used 44 videos for training.

The model was fine-tuned from the network proposed in~\cite{LWang-CVPR2015} using randomly initialized weights for new layers. We trained the network for 300k iterations, using the Stochastic Gradient Descent (SGD) optimization~\cite{Rumelhart-Neurocomputing1988} with a moment $\beta=0.9$ and a weight decay of 0.005. The size of each mini-batch is set 500. A base learning rate was initially set to $0.001$ and divided by 10 at every 50k iterations.

\section{Experimental Results} \label{section:experiments}

\subsection{Comparison with the State-of-the-Arts}

We compared the performance of our method (denoted by STCRF) with several state-of-the-art methods for salient object detection such as LC~\cite{Zhai-MM2006}, LD~\cite{Liu-PAMI2011}, LGFOGR~\cite{Wang-TIP2015}, LRSD~\cite{Xue-ICASSP2012}, RST~\cite{Nghia-PSIVT2015}, SAG~\cite{Wang-CVPR2015}, SEG~\cite{Rahtu-ECCV2010}, STS~\cite{Zhou-CVPR2014}, DCL~\cite{Li-CVPR2016}, DHS~\cite{Liu-CVPR2016}, DS~\cite{Xi-TIP2015}, DSS~\cite{Hou-CVPR2017}, ELD~\cite{Lee-CVPR2016}, MDF~\cite{Li-CVPR2015}, and RFCN~\cite{Wang-ECCV2016}. Compared methods are classified in Table \ref{tab:compared_method}. We remark that we run their original codes provided by the authors with the recommended parameter settings for obtaining results. We also note that we applied the methods developed for the still image to videos frame-by-frame.

\begin{table}[t]
\centering
\caption{Compared state-of-the-art methods and classification.}
\label{tab:compared_method}
\resizebox{\linewidth}{!}{%
\begin{tabular}{l|cc}
\toprule
\textbf{Target} & \textbf{Hand-crafted feature} & \textbf{Deep feature} \\ \midrule
\textbf{Video}  & \begin{tabular}[c]{@{}l@{}}LC\cite{Zhai-MM2006},LD\cite{Liu-PAMI2011}, LGFOGR\cite{Wang-TIP2015}, \\ LRSD\cite{Xue-ICASSP2012}, RST\cite{Nghia-PSIVT2015}, SAG\cite{Wang-CVPR2015}, \\ SEG\cite{Rahtu-ECCV2010}, STS\cite{Zhou-CVPR2014}\end{tabular} & None \\ \midrule
\textbf{Image}  & None & \begin{tabular}[c]{@{}l@{}}DCL\cite{Li-CVPR2016}, DHS\cite{Liu-CVPR2016}, DS\cite{Xi-TIP2015}, \\ DSS\cite{Hou-CVPR2017}, ELD\cite{Lee-CVPR2016}, MDF\cite{Li-CVPR2015}, \\ RFCN\cite{Wang-ECCV2016}
\end{tabular} \\ \bottomrule
\end{tabular}
}
\end{table}

\begin{figure*}[p]
    \centering
    \includegraphics[width=1\textwidth]{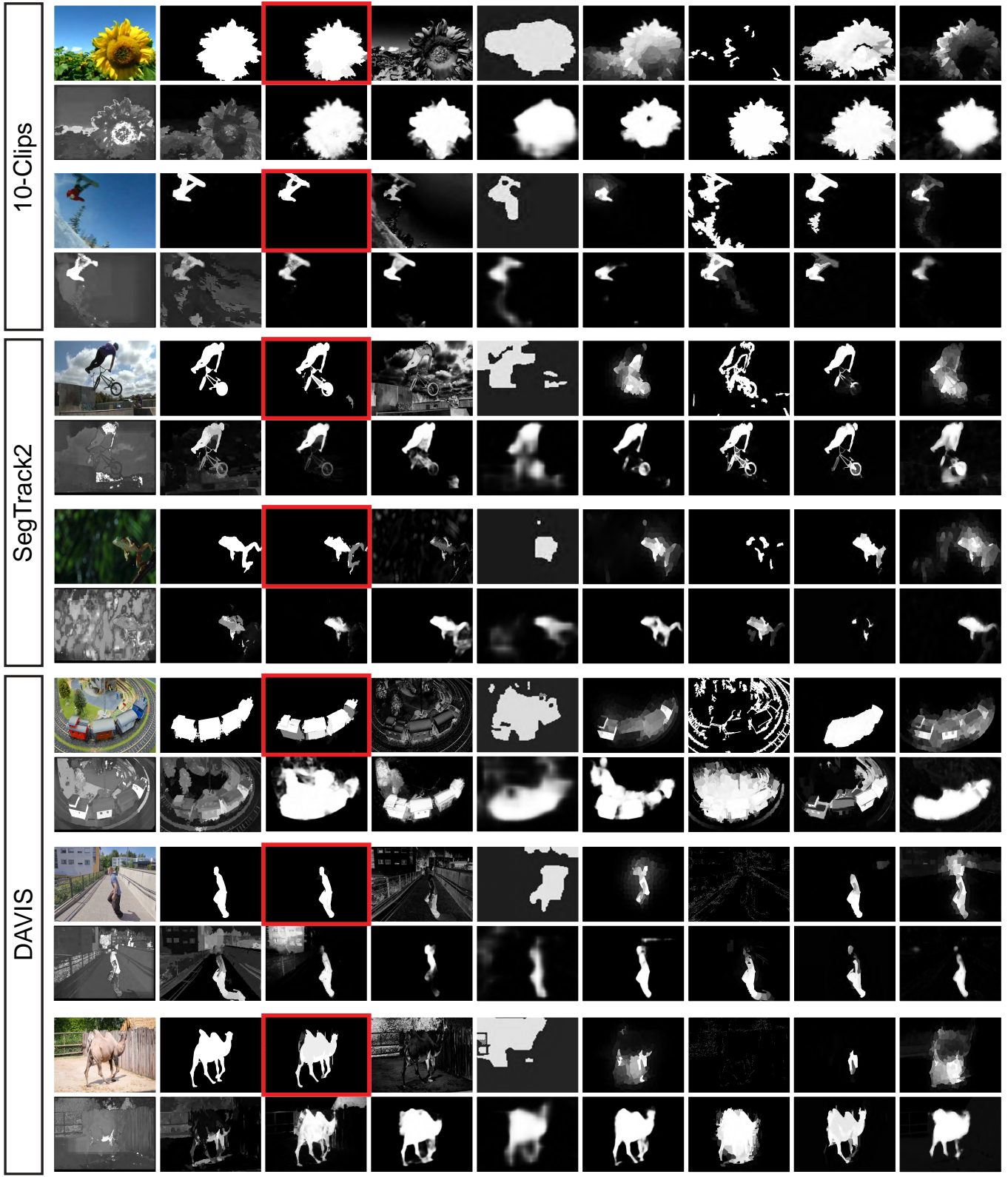}
    
    \caption{Visual comparison of our method against the state-of-the-art methods. From top-left to bottom-right, original video frame and ground-truth are followed by outputs obtained using our method (STCRF), LC\cite{Zhai-MM2006},LD\cite{Liu-PAMI2011}, LGFOGR\cite{Wang-TIP2015}, LRSD\cite{Xue-ICASSP2012}, RST\cite{Nghia-PSIVT2015}, SAG\cite{Wang-CVPR2015}, SEG\cite{Rahtu-ECCV2010}, STS\cite{Zhou-CVPR2014}, DCL\cite{Li-CVPR2016}, DHS\cite{Liu-CVPR2016}, DS\cite{Xi-TIP2015}, DSS\cite{Hou-CVPR2017}, ELD\cite{Lee-CVPR2016}, MDF\cite{Li-CVPR2015}, and RFCN\cite{Wang-ECCV2016}, in this order. Our method surrounded with red rectangles achieves the best results.}
\label{img:visual_comparison}
\end{figure*}

Figure \ref{img:visual_comparison} shows examples of obtained results. Qualitative evaluation confirms that our method produces the best results on each dataset. Our method can handle complex foreground and background with different details, giving accurate and uniform saliency assignment. In particular, object boundaries are clearly kept with less noise, compared with the other methods.

\begin{figure*}[t]
    \centering
        \footnotesize
    \begin{tabularx}{\textwidth}{*{3}{X}}
        \includegraphics[width=1\linewidth]{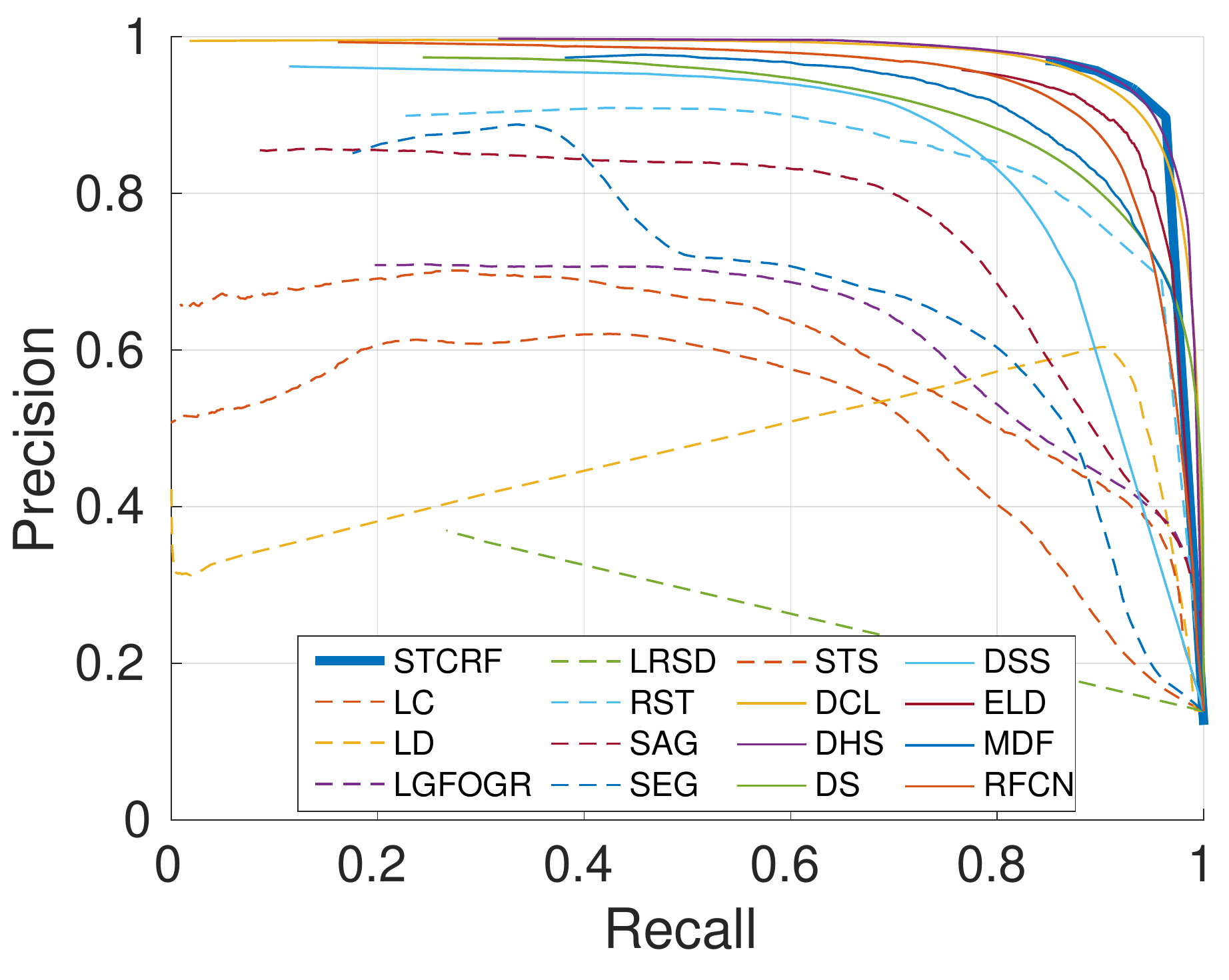} &
        \includegraphics[width=1\linewidth]{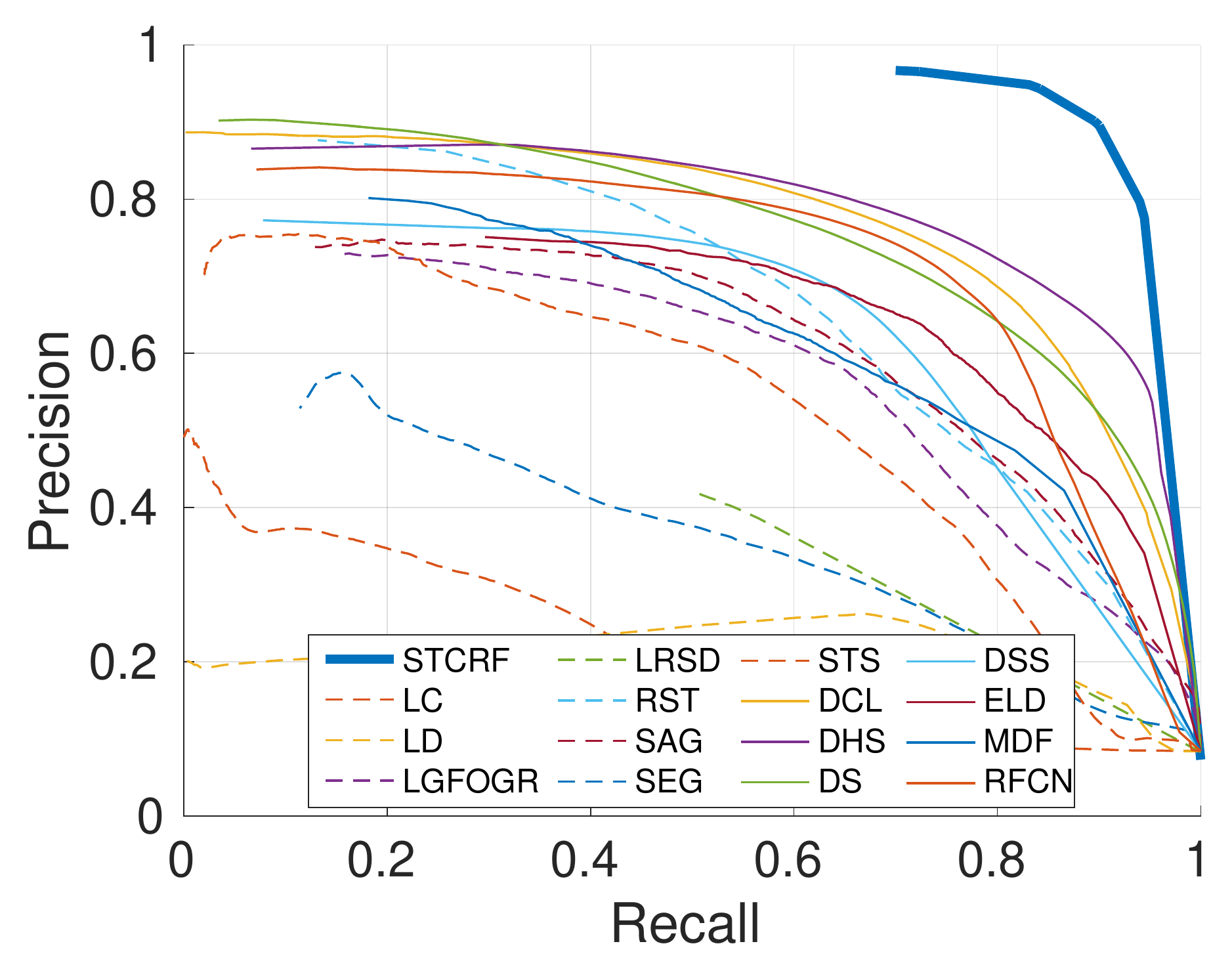} & 
        \includegraphics[width=1\linewidth]{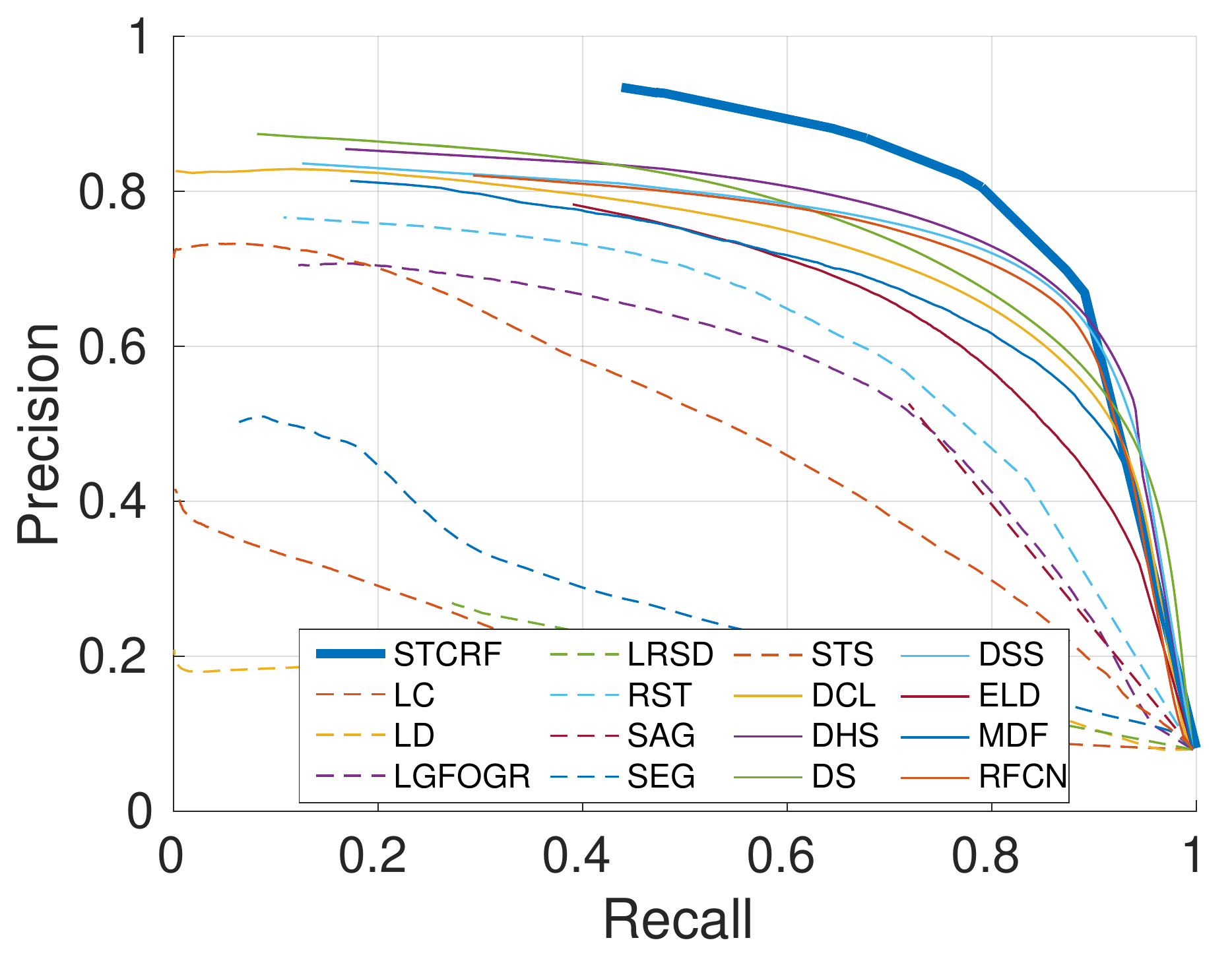} \\
        \centering (a) 10-Clips Dataset &
        \centering (b) SegTrack2 Dataset &
        \centering (c) DAVIS Dataset \\
    \end{tabularx}
    \caption{Quantitative comparison of precision-recall curve with state-of-the-art methods under different thresholds. Our method is denoted by STCRF (\textcolor[rgb]{0,0,1}{thick blue}).}
\label{fig:PRC}
\end{figure*}

\begin{figure*}[t]
    \centering
        \footnotesize
    \begin{tabularx}{\textwidth}{*{3}{X}}
        \includegraphics[width=1\linewidth]{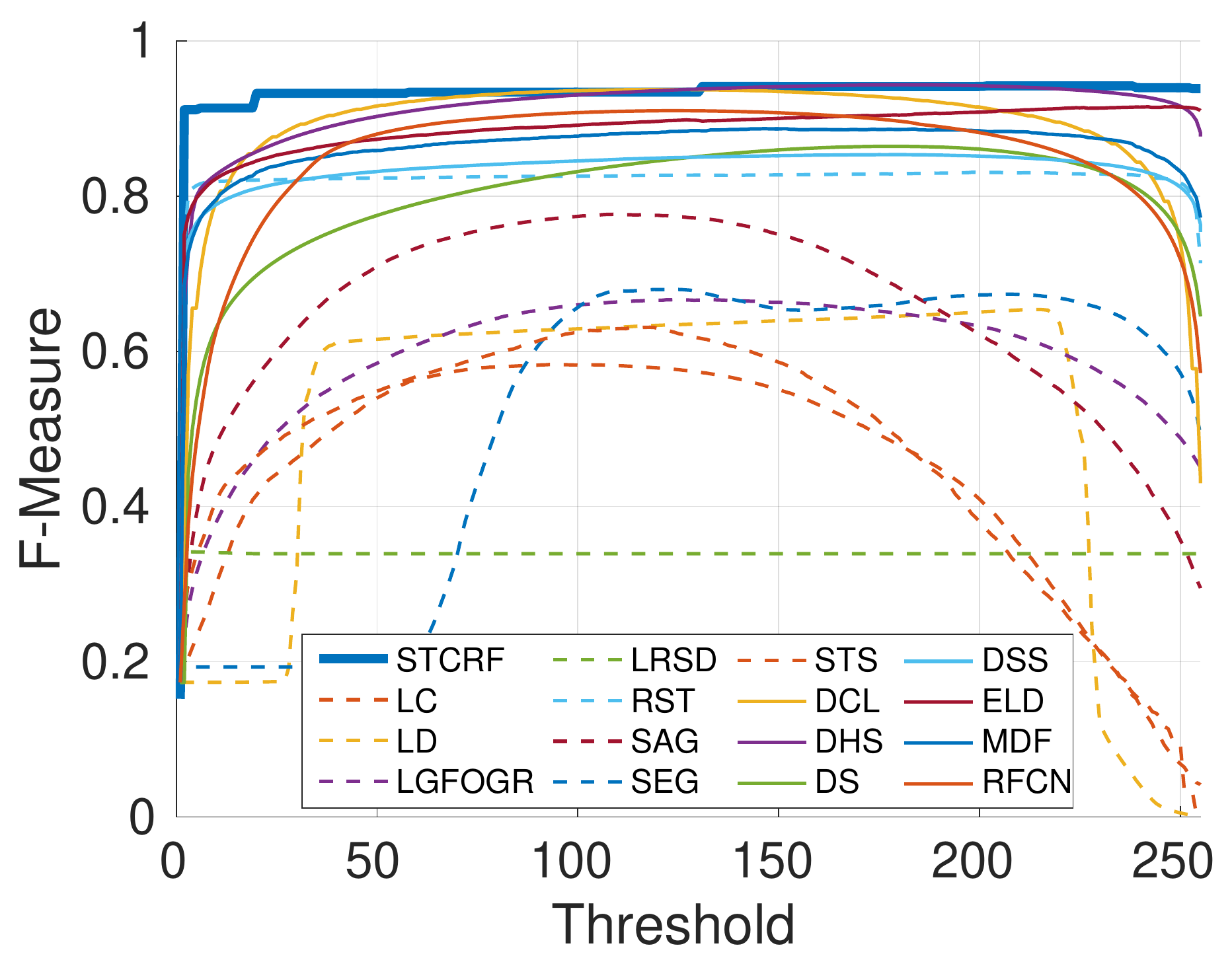} &
        \includegraphics[width=1\linewidth]{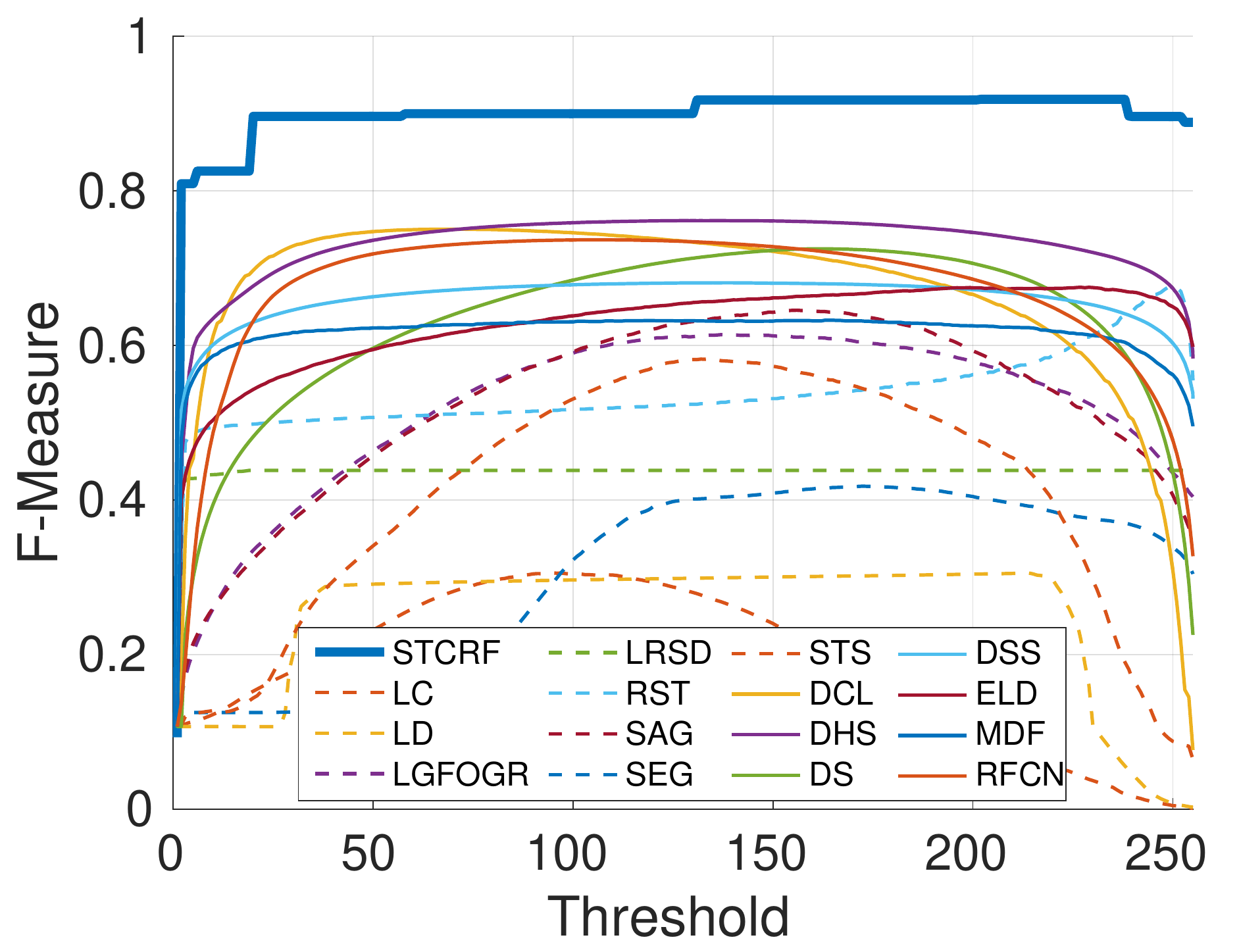} &
        \includegraphics[width=1\linewidth]{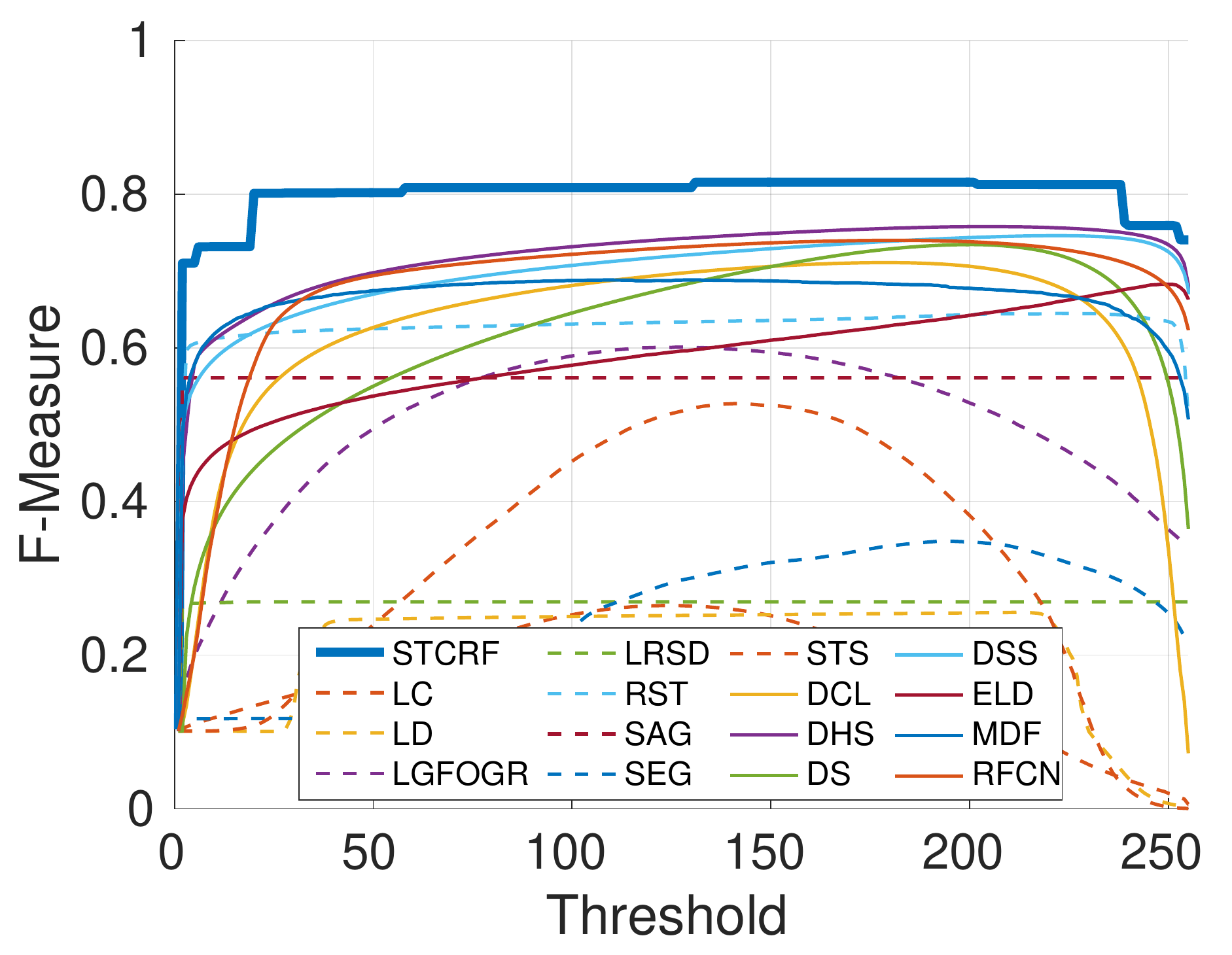} \\
        \centering (a) 10-Clips Dataset &
        \centering (b) SegTrack2 Dataset &
        \centering (c) DAVIS Dataset \\
    \end{tabularx}
    \caption{Quantitative comparison of F-measure with state-of-the-art methods under different thresholds. Our method is denoted by STCRF (\textcolor[rgb]{0,0,1}{thick blue}).}
\label{fig:F_curve}
\end{figure*}

To quantitatively evaluate the obtained results, we first computed PRC and F-measure curves, which are shown in Figs.~\ref{fig:PRC} and \ref{fig:F_curve}. 

It can be seen that our method achieves the highest precision in almost the entire recall ranges on all the datasets. Especially on the two most challenging datasets (i.e., SegTrack2 and DAVIS), the performance gains of our method against the other methods are more remarkable (results with higher recall values are less important because achieving higher recall values is easy). When compared with the second best method, i.e., DHS, we see that (1) both the methods have comparable results on 10-Clips dataset, that (2) our method is significantly better than DHS on SegTrack2 dataset, and that (3) on DAVIS dataset, the precision of our method is larger than that of DHS when recall values are small (higher binarization thresholds) while it is smaller for large recall values (lower binarization thresholds).  Salient object detection at higher thresholds is more practical and effective than that at lower thresholds because with low thresholds, more pixels are segmented regardless of salient objects or background.

F-measure indicates that our method significantly outperforms the other methods at every threshold on all the datasets. Since the 10-Clips dataset is easiest among the three datasets, any methods can achieve good results while the other two datasets are challenging, meaning that the effectiveness of methods becomes discriminative. Indeed, compared with the second best method (DHS), our method is comparable on the 10-Clips dataset and significantly better on the other datasets.

Table \ref{tab:comparison} illustrates the evaluations in terms of F-Adap, F-Max, and MAE. Our proposed method achieves the best performance under all the metrics on all the datasets. 
In particular, the outperformance of our method even against the second best method (DHS) is significant on SegTrack2 and DAVIS datasets. 

\begin{table}[t]
\centering
\caption{The wall-clock time average for each frame.}
\label{tab:time}
\begin{tabular}{l|cccc}
\toprule
\textbf{Method} & \textbf{Code} & \textbf{Platform} & \textbf{Time (seconds)} & \textbf{FPS} \\ \midrule
\textbf{STCRF} & Matlab &  CPU+GPU & \textbf{4.596}& \textbf{0.218} \\ 
\textbf{STCRF-full} & Matlab & CPU+GPU & \textbf{10.300}  & \textbf{0.097} \\ 

LGFOGR~\cite{Wang-TIP2015} & Matlab & CPU & 16.096 & 0.062  \\
RST~\cite{Nghia-PSIVT2015} & Matlab & CPU & 19.903 & 0.050  \\
SAG~\cite{Wang-CVPR2015} & Matlab & CPU & 17.613 & 0.057  \\
STS~\cite{Zhou-CVPR2014} & Matlab & CPU & 10.924 & 0.092  \\
MDF~\cite{Li-CVPR2015} & Matlab & CPU+GPU & 12.300 & 0.081 \\ \midrule

LD~\cite{Liu-PAMI2011} & Matlab & CPU & 8.318 & 0.120  \\
LRSD~\cite{Xue-ICASSP2012} & Matlab & CPU & 0.755 & 1.325  \\
SEG~\cite{Rahtu-ECCV2010} & Matlab & CPU &  4.856 & 0.206 \\
RFCN~\cite{Wang-ECCV2016} & Matlab & CPU+GPU & 1.840 & 0.543 \\ \midrule

LC~\cite{Zhai-MM2006} & C/C++ & CPU & 0.131 & 7.634  \\
DCL~\cite{Li-CVPR2016} & C/C++ & GPU &  0.183 & 5.464 \\
DHS~\cite{Liu-CVPR2016} & C/C++ & GPU & 0.080 & 12.500 \\
DS~\cite{Xi-TIP2015} & C/C++ & GPU & 0.109 & 9.174  \\
DSS~\cite{Hou-CVPR2017} & C/C++ & GPU & 0.178 & 5.618 \\
ELD~\cite{Lee-CVPR2016} & C/C++ & GPU & 2.030 & 0.493 \\

\bottomrule
\end{tabular}
\end{table}

\begin{table}[t]
\centering
\caption{The wall-clock time average of each step for each frame in the proposed method. Bottlenecks are shown in \textcolor[rgb]{1,0,0}{red}. (The local feature extraction stream and the global feature extraction stream run in parallel.)}
\label{tab:time_step}
\resizebox{1\linewidth}{!}{%
\begin{tabular}{l|l|c}
\toprule
\textbf{Part} & \textbf{Small step} & \textbf{Time (seconds)} \\ \midrule

Optical flow~\cite{Weinzaepfel-ICCV2013} &  & \textcolor[rgb]{1,0,0}{1.265}  \\ \midrule
Video segmentation~\cite{yu-ICCV2015} & & \textcolor[rgb]{1,0,0}{4.439} \\ \midrule
STD feature extraction & &  \\ 
& Region-based feature extraction & \textcolor[rgb]{1,0,0}{2.323} \\
& Local feature computation & 0.383 \\ 
& Global feature extraction & 0.027 \\
& (sub-total) &  (2.706)    \\ \midrule
Saliency computation & &  \\
& Unary potential prediction & 0.180 \\
& Binary potential computation & \textcolor[rgb]{1,0,0}{1.461} \\
& Saliency inference & 0.249 \\  
& (sub-total) &  (1.890)    \\ \midrule 
\textbf{Total} &  &  \textbf{10.300} \\

\bottomrule
\end{tabular}
}
\end{table}

\begin{table*}[t]
\footnotesize
\centering
\caption{Quantitative comparison with state-of-the-art methods, using F-measure (F-Adap and F-Max) (higher is better) and Mean Absolute Errors (MAE) (smaller is better). The best and the second best results are shown in \textcolor[rgb]{0,0,1}{\textbf{blue}} and \textcolor[rgb]{0,0.7,0}{\textbf{green}}, respectively. Our method (STCRF) marked in \textbf{bold} is followed by methods for videos and those for still images.}
\label{tab:comparison}
\begin{tabular}{lccccccccccc}
\toprule

\textbf{Dataset} & \multicolumn{3}{c}{\textbf{10-Clips}} & \multicolumn{1}{c}{\textbf{}} & \multicolumn{3}{c}{\textbf{SegTrack2}} & \multicolumn{1}{c}{\textbf{}} & \multicolumn{3}{c}{\textbf{DAVIS}} \\ \cline{2-4} \cline{6-8} \cline{10-12} 
\textbf{Metric} & \multicolumn{1}{c}{\textbf{F-Adap$\Uparrow$}} & \multicolumn{1}{c}{\textbf{F-Max$\Uparrow$}} & \multicolumn{1}{c}{\textbf{MAE$\Downarrow$}} & \multicolumn{1}{c}{\textbf{}} & \multicolumn{1}{c}{\textbf{F-Adap$\Uparrow$}} & \multicolumn{1}{c}{\textbf{F-Max$\Uparrow$}} & \multicolumn{1}{c}{\textbf{MAE$\Downarrow$}} & \multicolumn{1}{c}{\textbf{}} & \multicolumn{1}{c}{\textbf{F-Adap$\Uparrow$}} & \multicolumn{1}{c}{\textbf{F-Max$\Uparrow$}} & \multicolumn{1}{c}{\textbf{MAE$\Downarrow$}} \\ \midrule

\textbf{STCRF}        & \textcolor[rgb]{0,0,1}{\textbf{0.936}} & \textcolor[rgb]{0,0.7,0}{\textbf{0.942}} & \textcolor[rgb]{0,0,1}{\textbf{0.016}} &  & \textcolor[rgb]{0,0,1}{\textbf{0.899}} & \textcolor[rgb]{0,0,1}{\textbf{0.919}} & \textcolor[rgb]{0,0,1}{\textbf{0.014}} &  & \textcolor[rgb]{0,0,1}{\textbf{0.803}} & \textcolor[rgb]{0,0,1}{\textbf{0.816}} & \textcolor[rgb]{0,0,1}{\textbf{0.033}} \\ \midrule

LC~\cite{Zhai-MM2006}      & 0.577 & 0.583 & 0.166 &  & 0.244 & 0.306 & 0.173 &  & 0.161 & 0.209 & 0.203 \\
LD~\cite{Liu-PAMI2011}     & 0.637 & 0.654 & 0.197 &  & 0.286 & 0.305 & 0.281 &  & 0.242 & 0.246 & 0.302 \\
LGFOGR~\cite{Wang-TIP2015} & 0.629 & 0.667 & 0.207 &  & 0.500 & 0.614 & 0.117 &  & 0.555 & 0.614 & 0.100 \\
LRSD~\cite{Xue-ICASSP2012} & 0.339 & 0.342 & 0.164 &  & 0.438 & 0.438 & 0.102 &  & 0.228 & 0.229 & 0.109 \\
RST~\cite{Nghia-PSIVT2015} & 0.827 & 0.831 & 0.055 &  & 0.510 & 0.677 & 0.125 &  & 0.607 & 0.628 & 0.081 \\
SAG~\cite{Wang-CVPR2015}   & 0.755 & 0.777 & 0.117 &  & 0.504 & 0.646 & 0.106 &  & 0.487 & 0.537 & 0.105 \\
SEG~\cite{Rahtu-ECCV2010}  & 0.687 & 0.680 & 0.298 &  & 0.388 & 0.418 & 0.321 &  & 0.313 & 0.345 & 0.316 \\
STS~\cite{Zhou-CVPR2014}   & 0.591 & 0.631 & 0.177 &  & 0.471 & 0.583 & 0.147 &  & 0.362 & 0.489 & 0.194 \\ \midrule
DCL~\cite{Li-CVPR2016}     & \textcolor[rgb]{0,0.7,0}{\textbf{0.935}} & 0.937 & 0.031 &  & \textcolor[rgb]{0,0.7,0}{\textbf{0.734}} & 0.750 & 0.060 &  & 0.630 & 0.673 & 0.075 \\
DHS~\cite{Liu-CVPR2016} & 0.923 & \textcolor[rgb]{0,0,1}{\textbf{0.947}} & \textcolor[rgb]{0,0.7,0}{\textbf{0.022}} &  & 0.733 & \textcolor[rgb]{0,0.7,0}{\textbf{0.762}} & \textcolor[rgb]{0,0.7,0}{\textbf{0.050}} &  & \textcolor[rgb]{0,0.7,0}{\textbf{0.738}} & \textcolor[rgb]{0,0.7,0}{\textbf{0.777}} & \textcolor[rgb]{0,0.7,0}{\textbf{0.040}} \\
DS~\cite{Xi-TIP2015}       & 0.832 & 0.864 & 0.050 &  & 0.636 & 0.725 & 0.069 &  & 0.610 & 0.741 & 0.083 \\
DSS~\cite{Hou-CVPR2017}    & 0.838 & 0.853 & 0.049 &  & 0.662 & 0.681 & 0.054 &  & 0.693 & 0.758 & 0.047 \\
ELD~\cite{Lee-CVPR2016}    & 0.893 & 0.915 & 0.023 &  & 0.611 & 0.675 & 0.065 &  & 0.571 & 0.686 & 0.076 \\
MDF~\cite{Li-CVPR2015}     & 0.884 & 0.887 & 0.041 &  & 0.627 & 0.633 & 0.077 &  & 0.670 & 0.673 & 0.067 \\
RFCN~\cite{Wang-ECCV2016}  & 0.901 & 0.910 & 0.046 &  & 0.716 & 0.737 & 0.062 &  & 0.694 & 0.725 & 0.070 \\ 
\bottomrule
\end{tabular}
\end{table*} 

\subsection{Computational Efficiency}

We further evaluated the computational time of all the methods. 
We compared the running-time average of our method with that of the other methods. 
The wall-clock time average for each frame in our method and the compared methods is given in Table \ref{tab:time}.
Our methods are denoted by STCRF for the pipeline without counting optical flow computation and video segmentation, and by STCRF-full for the full pipeline. 
We note that all videos were resized to the resolution of $352 \times 288$ for the fair comparison. 
Performances of all the methods are compared based on the implementations in C/C++ and Matlab. 
We classify all the methods into three categories: Matlab-region-based methods, Matlab-pixel-based methods, and C/C++ based methods.

Since it is obvious that codes implemented in C/C++ run faster than those in Matlab, we cannot directly compare the run-time of all the methods.
However, we see that our method runs in the competitive speed with the others.
Indeed, our method is fastest among the Matlab-region-based methods. 
We remark that Matlab-region-based methods run more slowly than Matlab-pixel-based ones because treating regions individually in a sequential manner and then integrating results are time-consuming. 

It can be seen in Table \ref{tab:time} that in our method, the time required for computing optical flow and video segmentation is a bottleneck: it takes 5.704 ($=10.300-4.596$) seconds per frame.  To identify bottleneck steps in our pipeline, we broke down running-time into individual steps in our pipeline (see Table \ref{tab:time_step}).
We note that in our pipeline, the step of region-based feature extraction followed by local feature computation, and the step of global feature extraction run in parallel.
Table \ref{tab:time_step} indicates that region-based feature extraction and binary potential computation are also bottlenecks.  
Because the bottleneck steps except for region-based feature extraction are implemented in Matlab,   
re-implementing such steps in C/C++ and using Cuda for parallel processing for regions will improve the speed of our method.
We note that speed-up of the computational time for salient object detection is not the scope of this paper.

\subsection{Detailed Analysis of the Proposed Method}\label{section:validation}

To demonstrate the effectiveness of utilizing local and global features, utilizing spatiotemporal information in computing the saliency map, and the effectiveness of multi-level analysis, we performed experiments under controlled settings and compared results.

\subsubsection{Effectiveness of Combination of Local and Global Features}

To evaluate the effectiveness of combining local and global features, we compared results using STD features with those using local features alone, which is illustrated in Table \ref{tab:feature}.

We see that the combination of local and global features brings more gains than using only local features.
This can be explained as follows. Local features exploit the meaning of an object in term of saliency but only in a local context, while global features can model a global context in the whole video block. Thus, STD features are more powerful. 
We remark that we also present results using RGB features just to confirm that the deep feature outperforms RGB features.

\begin{table*}[t]
\centering
\caption{Comparison of STD features and local features. The best results are shown in \textcolor[rgb]{0,0,1}{blue} (higher is better for F-Adap and F-Max, and lower is better for MAE). 
}
\label{tab:feature}
\begin{tabular}{lccccccccccc}
\toprule

\multicolumn{1}{c}{\multirow{2}{*}{\textbf{Used feature}}} & \multicolumn{3}{c}{\textbf{10-Clips}} & & \multicolumn{3}{c}{\textbf{SegTrack2}} & & \multicolumn{3}{c}{\textbf{DAVIS}} \\ \cline{2-4}\cline{6-8}\cline{10-12}
 
 &  
\multicolumn{1}{c}{\textbf{F-Adap $\Uparrow$}} & \multicolumn{1}{c}{\textbf{F-Max $\Uparrow$}} & \multicolumn{1}{c}{\textbf{MAE $\Downarrow$}} & &
\multicolumn{1}{c}{\textbf{F-Adap $\Uparrow$}} & \multicolumn{1}{c}{\textbf{F-Max $\Uparrow$}} & \multicolumn{1}{c}{\textbf{MAE $\Downarrow$}} & &
\multicolumn{1}{c}{\textbf{F-Adap $\Uparrow$}} & \multicolumn{1}{c}{\textbf{F-Max $\Uparrow$}} & \multicolumn{1}{c}{\textbf{MAE $\Downarrow$}} \\ \midrule

\textbf{STD feature}         & \textcolor[rgb]{0,0,1}{\textbf{0.936}} & \textcolor[rgb]{0,0,1}{\textbf{0.942}} & \textcolor[rgb]{0,0,1}{\textbf{0.016}} & & \textcolor[rgb]{0,0,1}{\textbf{0.899}} & \textcolor[rgb]{0,0,1}{\textbf{0.919}} & \textcolor[rgb]{0,0,1}{\textbf{0.014}} & & \textcolor[rgb]{0,0,1}{\textbf{0.803}} & \textcolor[rgb]{0,0,1}{\textbf{0.816}} & \textcolor[rgb]{0,0,1}{\textbf{0.033}} \\
Local feature alone  & 0.683 & 0.727 & 0.079 & & 0.692 & 0.780 & 0.043 & & 0.648 & 0.744 & 0.067 \\
RGB feature    & 0.882 & 0.913 & 0.044 & & 0.366 & 0.454 & 0.080 & & 0.160 & 0.186 & 0.199 \\

\bottomrule
\end{tabular}
\end{table*}

\subsubsection{Effectiveness of Spatiotemporal Potential in STCRF}

\begin{table*}[t]
\centering
\caption{Comparison of different potentials in STCRF. The best results are shown in \textcolor[rgb]{0,0,1}{blue} (higher is better for F-Adap and F-Max, and lower is better for MAE). Our complete method are marked in \textbf{bold}.}
\label{tab:potential}
\resizebox{\textwidth}{!}{%
\begin{tabular}{l|cccccccccccccc}
\toprule

\multicolumn{1}{c|}{} & \multicolumn{1}{c}{} & \multicolumn{1}{c}{} & \multicolumn{1}{c}{} & \multicolumn{3}{c}{\textbf{10-Clips}} & & \multicolumn{3}{c}{\textbf{SegTrack2}} & & \multicolumn{3}{c}{\textbf{DAVIS}} \\ \cline{5-7} \cline{9-11} \cline{13-15}
 
\multicolumn{1}{c|}{\multirow{-2}{*}{\textbf{\begin{tabular}[c]{@{}c@{}}Setting \\ description \end{tabular}}}} & 
\multicolumn{1}{c}{\multirow{-2}{*}{\textbf{\begin{tabular}[c]{@{}c@{}} Unary\\ term \end{tabular}}}} &
\multicolumn{1}{c}{\multirow{-2}{*}{\textbf{\begin{tabular}[c]{@{}c@{}} Spatial\\ information \end{tabular}}}} &
\multicolumn{1}{c}{\multirow{-2}{*}{\textbf{\begin{tabular}[c]{@{}c@{}}Temporal\\ information \end{tabular}}}} &
\multicolumn{1}{c}{\textbf{F-Adap $\Uparrow$}} & \multicolumn{1}{c}{\textbf{F-Max $\Uparrow$}} & \multicolumn{1}{c}{\textbf{MAE $\Downarrow$}} & &
\multicolumn{1}{c}{\textbf{F-Adap $\Uparrow$}} & \multicolumn{1}{c}{\textbf{F-Max $\Uparrow$}} & \multicolumn{1}{c}{\textbf{MAE $\Downarrow$}} & &
\multicolumn{1}{c}{\textbf{F-Adap $\Uparrow$}} & \multicolumn{1}{c}{\textbf{F-Max $\Uparrow$}} & \multicolumn{1}{c}{\textbf{MAE $\Downarrow$}} \\ \midrule

\textbf{STP} & $\vee$ & $\vee$   & $\vee$   & \textcolor[rgb]{0,0,1}{\textbf{0.936}} & \textcolor[rgb]{0,0,1}{\textbf{0.942}} & \textcolor[rgb]{0,0,1}{\textbf{0.016}} & & \textcolor[rgb]{0,0,1}{\textbf{0.899}} & \textcolor[rgb]{0,0,1}{\textbf{0.919}} & \textcolor[rgb]{0,0,1}{\textbf{0.014}} & & \textcolor[rgb]{0,0,1}{\textbf{0.803}} & \textcolor[rgb]{0,0,1}{\textbf{0.816}} & \textcolor[rgb]{0,0,1}{\textbf{0.033}} \\
SP    & $\vee$ & $\times$ & $\vee$   & 0.930 & 0.941 & 0.018 & & 0.871 & 0.918 & 0.017 & & 0.759 & 0.814 & 0.038 \\
TP    & $\vee$ & $\vee$   & $\times$ & 0.930 & 0.940 & 0.019 & & 0.859 & 0.901 & 0.019 & & 0.750 & 0.805 & 0.039 \\
U       & $\vee$ & $\times$ & $\times$ & 0.876 & 0.940 & 0.044 & & 0.703 & 0.912 & 0.072 & & 0.537 & 0.804 & 0.169 \\
\bottomrule
\end{tabular}
}
\end{table*}

To demonstrate the effectiveness of utilizing spatiotemporal information into the energy function in STRCF, we performed experiments under four different controlled settings. 
We changed the binary term: setting $\theta_{\rm bt}=0$ to use spatial information alone (denoted by SP), 
setting $\theta_{\rm bs}=0$ to use temporal information alone (denoted by TP), and 
setting $\theta_{\rm bt}=\theta_{\rm bs}=0$ to use the unary term alone (denoted by U).
We compared the proposed (complete) method (denoted by STP) with these three baseline methods (cf. Table \ref{tab:potential}). 

Table \ref{tab:potential} indicates that STP exhibits the best performance on all the metrics on the three datasets. We see that using both spatial and temporal information effectively works and brings more gains than using spatial information alone or using temporal information alone. This suggests that our method captures spatial contexts in a frame and temporal information over frames to produce saliency maps.

\subsubsection{Effectiveness of Multiple-Scale Approach}

To demonstrate the effectiveness of our multiple-scale approach, we compared methods that use different numbers of scale levels in computing the saliency map.
More precisely, starting with only the coarsest scale level (level 1), we fused finer levels (levels 2, 3, 4) one by one to 
compute the saliency map. 
The methods are denoted by 1-level, 2-levels, 3-levels, and 4-levels (our complete method).

The results are illustrated in Table \ref{tab:multiscale}.  Table \ref{tab:multiscale} shows that the multiple-scale approach outperforms the single-scale approach. 
It also indicates that using more scales produces better results.  Indeed, as the number of scales in the saliency computation increases, we have more accurate results.
Table \ref{tab:multiscale} also shows that employing
4 scale levels seems to be sufficient because 3-levels and 4-levels have almost similar performances.

\begin{table*}[t]
\centering
\caption{Comparison of different numbers of scale levels in processing. The best results are shown in \textcolor[rgb]{0,0,1}{blue} (higher is better for F-Adap and F-Max, and lower is better for MAE). Our complete method is marked in \textbf{bold}.}
\label{tab:multiscale}
\begin{tabular}{lccccccccccc}
\toprule

\multicolumn{1}{c}{} & \multicolumn{3}{c}{\textbf{10-Clips}} & & \multicolumn{3}{c}{\textbf{SegTrack2}} & & \multicolumn{3}{c}{\textbf{DAVIS}} \\ \cline{2-4}\cline{6-8}\cline{10-12}
 
\multicolumn{1}{c}{\multirow{-2}{*}{\textbf{\begin{tabular}[c]{@{}c@{}}Setting \\ description \end{tabular}}}} &  
\multicolumn{1}{c}{\textbf{F-Adap $\Uparrow$}} & \multicolumn{1}{c}{\textbf{F-Max $\Uparrow$}} & \multicolumn{1}{c}{\textbf{MAE $\Downarrow$}} & &
\multicolumn{1}{c}{\textbf{F-Adap $\Uparrow$}} & \multicolumn{1}{c}{\textbf{F-Max $\Uparrow$}} & \multicolumn{1}{c}{\textbf{MAE $\Downarrow$}} & &
\multicolumn{1}{c}{\textbf{F-Adap $\Uparrow$}} & \multicolumn{1}{c}{\textbf{F-Max $\Uparrow$}} & \multicolumn{1}{c}{\textbf{MAE $\Downarrow$}} \\ \midrule

1-level   & 0.928 & 0.930 & 0.017 & & 0.880 & 0.889 & 0.016 & & 0.750 & 0.757 & 0.038 \\

2-levels  & 0.929 & 0.940 & 0.017 & & 0.876 & 0.908 & 0.015 & & 0.763 & 0.800 & 0.035 \\
3-levels  & 0.935 & 0.941 & \textcolor[rgb]{0,0,1}{\textbf{0.016}} & & \textcolor[rgb]{0,0,1}{\textbf{0.909}} & 0.912 & \textcolor[rgb]{0,0,1}{\textbf{0.014}} & & 0.798 & \textcolor[rgb]{0,0,1}{\textbf{0.816}} & \textcolor[rgb]{0,0,1}{\textbf{0.033}} \\

\textbf{4-levels}         & \textcolor[rgb]{0,0,1}{\textbf{0.936}} & \textcolor[rgb]{0,0,1}{\textbf{0.942}} & \textcolor[rgb]{0,0,1}{\textbf{0.016}} & & 0.899 & \textcolor[rgb]{0,0,1}{\textbf{0.919}} & \textcolor[rgb]{0,0,1}{\textbf{0.014}} & & \textcolor[rgb]{0,0,1}{\textbf{0.803}} & \textcolor[rgb]{0,0,1}{\textbf{0.816}} & \textcolor[rgb]{0,0,1}{\textbf{0.033}} \\

\bottomrule
\end{tabular}
\end{table*}

\subsubsection{Effective Length of Video Block}\label{section:ws}

\begin{table*}[t]
\centering
\caption{Comparison under different lengths of the video block.}
\label{tab:ws}
\begin{tabular}{ccccccccc}
\toprule

\multicolumn{1}{c}{} & \multicolumn{2}{c}{\textbf{10-Clips}} & & \multicolumn{2}{c}{\textbf{SegTrack2}} & & \multicolumn{2}{c}{\textbf{DAVIS}} \\ \cline{2-3}\cline{5-6}\cline{8-9}
 
\multicolumn{1}{c}{\multirow{-2}{*}{\textbf{\begin{tabular}[c]{@{}c@{}}Length of \\ video block \end{tabular}}}} &  
\multicolumn{1}{c}{\textbf{F-Adap $\Uparrow$}} & \multicolumn{1}{c}{\textbf{MAE $\Downarrow$}} & &
\multicolumn{1}{c}{\textbf{F-Adap $\Uparrow$}} & \multicolumn{1}{c}{\textbf{MAE $\Downarrow$}} & &
\multicolumn{1}{c}{\textbf{F-Adap $\Uparrow$}} & \multicolumn{1}{c}{\textbf{MAE $\Downarrow$}} \\ \midrule

$1\,\, (=2^0)$   & 0.934 & 0.017 & & 0.890 & 0.016 & & 0.791 & 0.035 \\
$2\,\, (=2^1)$   & 0.935 & 0.017 & & 0.892 & 0.015 & & 0.792 & 0.034 \\
$4\,\, (=2^2)$   & 0.935 & 0.017 & & 0.895 & 0.015 & & 0.794 & 0.034 \\
$8\,\, (=2^3)$   & 0.936 & 0.017 & & 0.897 & 0.015 & & 0.799 & 0.033 \\
\textbf{$16\,\, (=2^4)$}   & \textbf{0.936} & \textbf{0.016} & & \textbf{0.899} & \textbf{0.014} & & 0\textbf{.803} & \textbf{0.033} \\
$32\,\, (=2^5)$   & 0.936 & 0.016 & & 0.899 & 0.014 & & 0.803 & 0.032 \\
$64\,\, (=2^6)$   & 0.936 & 0.016 & & 0.899 & 0.014 & & 0.803 & 0.032 \\

\bottomrule
\end{tabular}
\end{table*}

We investigated the effectiveness of the size of the video block to feed to STCRF by changing the window size from 1 to 64 by twice: $1, 2, 2^2,\ldots, 2^6$ (cf. Table \ref{tab:ws}). 

Table \ref{tab:ws} shows that as the window size becomes larger, we have more accurate results. However, the improvement in accuracy is saturated around a size of 16. On the other hand, the processing time for a larger window size is slower because the size of the graphical model becomes larger. To balance the performance between accuracy and the processing time, we observe that the appropriate window size of the video block is 16.


\section{Application to Video Object Segmentation} \label{section:application}

\begin{figure*}[t]
    \centering
        \includegraphics[width=0.75\linewidth]{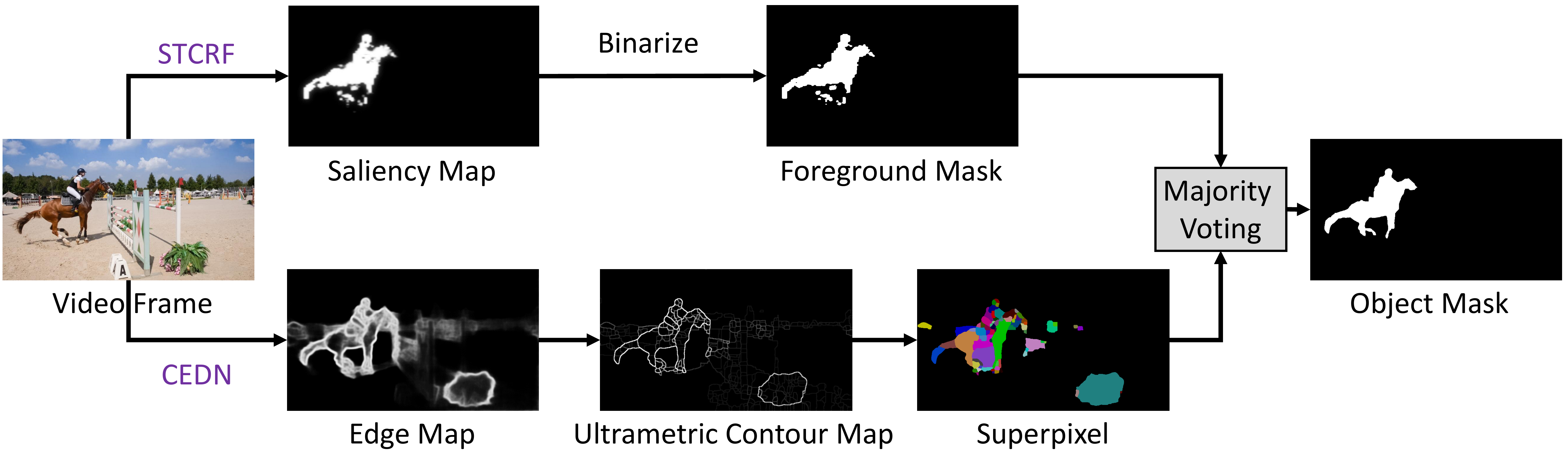}
    \caption{Boundary snapping~\cite{Caelles-CVPR2017} based video object segmentation framework using the saliency map.}
    \label{fig:VOS}
\end{figure*}

\begin{table*}[t]
\centering
\caption{Quantitative comparison with state-of-the-art video object segmentation methods on the DAVIS dataset, using region similarity, contour accuracy, and overall performance metrics. The best three results are shown in \textcolor[rgb]{0,0,1}{\textbf{blue}}, \textcolor[rgb]{0,0.7,0}{\textbf{green}}, and \textcolor[rgb]{1,0,0}{\textbf{red}}, respectively. Our method, denoted by STCRF*, is marked in \textbf{bold}.} 
\label{tab:vos}
\begin{tabular}{lccccccccc}
\toprule
\multicolumn{1}{c}{\multirow{2}{*}{\textbf{Methods}}} & \multicolumn{3}{c}{\textbf{Region similarity $(\mathcal{J})$}} & \textbf{} & \multicolumn{3}{c}{\textbf{Contour accuracy $(\mathcal{F})$}} & \textbf{} & \multicolumn{1}{c}{\textbf{Overall performance $(\mathcal{O})$}} \\ \cline{2-4} \cline{6-8} \cline{10-10} 

 & \textbf{Mean$\Uparrow$} & \textbf{Recall$\Uparrow$} & \textbf{Decay$\Downarrow$} & \textbf{} & \textbf{Mean$\Uparrow$} & \textbf{Recall$\Uparrow$} & \textbf{Decay$\Downarrow$} & \textbf{} & \textbf{\textbf{Mean$\Uparrow$}} \\ \midrule

\textbf{STCRF*}             & \textcolor[rgb]{0,0,1}{\textbf{0.714}} & \textcolor[rgb]{0,0,1}{\textbf{0.851}} & \textcolor[rgb]{0,0,1}{\textbf{-0.019}} &  & \textcolor[rgb]{0,0,1}{\textbf{0.674}} & \textcolor[rgb]{0,0,1}{\textbf{0.790}} & \textcolor[rgb]{0,0.7,0}{\textbf{-0.019}} &  & \textcolor[rgb]{0,0,1}{\textbf{0.694}} \\ 
DHS*~\cite{Liu-CVPR2016}      & \textcolor[rgb]{0,0.7,0}{\textbf{0.701}} & \textcolor[rgb]{0,0.7,0}{\textbf{0.840}} & 0.032 &  & \textcolor[rgb]{0,0.7,0}{\textbf{0.656}} & \textcolor[rgb]{0,0.7,0}{\textbf{0.779}} & 0.036 &  & \textcolor[rgb]{0,0.7,0}{\textbf{0.679}} \\ 
\midrule
ACO~\cite{Jang-CVPR2016}        & 0.503 & 0.572 & \textcolor[rgb]{0,0.7,0}{\textbf{-0.006}} &  & 0.467 & 0.494 & \textcolor[rgb]{0,0,1}{\textbf{-0.022}} &  & 0.485 \\ 
CVOS~\cite{Taylor-CVPR2015} & 0.482 & 0.540 & 0.105 && 0.447 & 0.526 & 0.117 && 0.465 \\ 
FST~\cite{Papazoglou-ICCV2013} & \textcolor[rgb]{1,0,0}{\textbf{0.558}} & \textcolor[rgb]{1,0,0}{\textbf{0.649}} & \textcolor[rgb]{1,0,0}{\textbf{0.000}} && 0.511 & 0.516 & \textcolor[rgb]{1,0,0}{\textbf{0.029}} && 0.535 \\ 
KEY~\cite{Lee-ICCV2011} & 0.498 & 0.591 & 0.141 && 0.427 & 0.375 & 0.106 && 0.463 \\ 
MSG~\cite{Ochs-ICCV2011} & 0.533 & 0.626 & 0.024  && 0.508 & \textcolor[rgb]{1,0,0}{\textbf{0.600}} & 0.051  && 0.521 \\ 
NLC~\cite{Faktor-BMVC2014} & 0.552 & 0.558 & 0.126  && \textcolor[rgb]{1,0,0}{\textbf{0.523}} & 0.519 & 0.114  && \textcolor[rgb]{1,0,0}{\textbf{0.537}} \\ 
TRC~\cite{Fragkiadaki-CVPR2012} & 0.473 & 0.493 & 0.083 && 0.441 & 0.436 & 0.129 && 0.457 \\ 
\bottomrule
\end{tabular}
\end{table*}

\begin{figure*}[h!]
    \centering
    \includegraphics[width=1\textwidth]{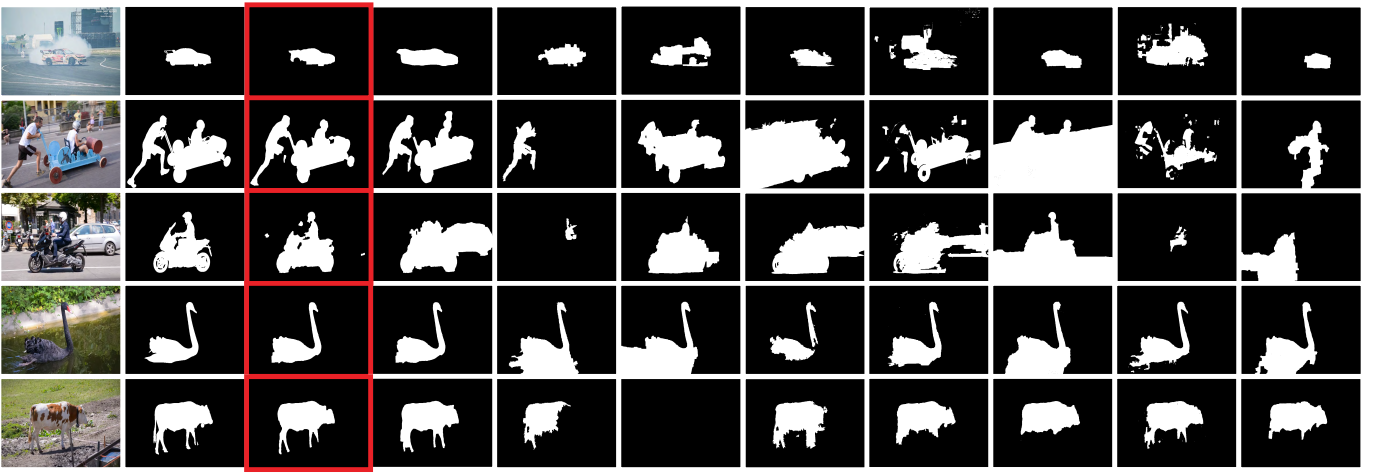}
    \caption{Visual comparison of our method against the state-of-the-art video object segmentation methods. From left to right, original video frame and ground-truth are followed by outputs obtained using our method (STCRF*), DHS*\cite{Liu-CVPR2016}, ACO\cite{Jang-CVPR2016}, CVOS\cite{Taylor-CVPR2015}, FST\cite{Papazoglou-ICCV2013}, KEY\cite{Lee-ICCV2011}, MSG\cite{Ochs-ICCV2011}, NLC\cite{Faktor-BMVC2014}, and TRC\cite{Fragkiadaki-CVPR2012}, 
    in this order. Our STCRF surrounded with red rectangles achieves the best results.}
\label{img:visual_comparison_vos}
\end{figure*}

Video object segmentation (VOS) is a binary labeling problem aiming to separate foreground objects from the background of a video~\cite{Perazzi-CVPR2016}. On the other hand, salient object detection (SOD) aims to detect and segment salient objects in natural scenes. Although VOS and SOD are different tasks, SOD methods are beneficial for VOS when salient objects are foreground objects in scenes. In this section, we demonstrate the applicability of our proposed method to VOS.

Figure \ref{fig:VOS} illustrates the framework for VOS using the saliency map. In one pass, the output saliency map is binarized using the adaptive threshold mentioned in Section \ref{section:metrics} to obtain the foreground mask. In the other pass, we implemented the object segmentation method based on boundary snapping~\cite{Caelles-CVPR2017}. We first detect contours of foreground objects using CEDN~\cite{Yang-CVPR2016} and then apply the combinatorial grouping method~\cite{Jordi-PAMI2017} to compute the Ultrametric Contour Map (UCM) ~\cite{Jordi-PAMI2017}, which presents hierarchical segmentation. Superpixels are aligned by binarizing the UCM using a threshold $\tau=0.3$. From the foreground mask and superpixels, we perform the majority voting to segment foreground objects. 

VOS methods are classified into two groups: one that is requiring the initial object mask at the first frame, and the other that is not. In the DAVIS Benchmark~\cite{Perazzi-CVPR2016}, the former group is called semi-supervised while the latter one is unsupervised. 
Since an initial object mask becomes a strong prior for accurately segmenting objects in subsequent frames, we chose most recent unsupervised methods for the fair comparison. We compared our method with the state-of-the-art saliency method (DHS~\cite{Liu-CVPR2016}), and most recent unsupervised VOS methods: ACO~\cite{Jang-CVPR2016}, CVOS~\cite{Taylor-CVPR2015}, FST~\cite{Papazoglou-ICCV2013}, KEY~\cite{Lee-ICCV2011}, MSG~\cite{Ochs-ICCV2011},  NLC~\cite{Faktor-BMVC2014}, and TRC~\cite{Fragkiadaki-CVPR2012}.
We remark that two SOD methods (i.e., our method and DHS) segment objects using the framework in Fig. \ref{fig:VOS}. We denote their by STCRF* and DHS* individually.

We tested all the methods on the DAVIS dataset~\cite{Perazzi-CVPR2016}, the newest dataset for VOS, and evaluated results using measures in the 2017 DAVIS Challenge~\cite{Jordi-2017} (i.e., region similarity $\mathcal{J}$, contour accuracy $\mathcal{F}$, and overall performance $\mathcal{O}$). For a given error measure, we computed three different statistics as in~\cite{Perazzi-CVPR2016}.  They are the mean error, the object recall (measuring the fraction of sequences scoring higher than a threshold $\tau=0.5$), and the decay (quantifying the performance loss (or gain) over time). Note that we used the results in the DAVIS Benchmark~\cite{Perazzi-CVPR2016}\footnote{\href{http://davischallenge.org/soa\_compare.html}{http://davischallenge.org/soa\_compare.html}}
for the compared state-of-the-art VOS techniques.
We also note that we run the source code of ACO~\cite{Jang-CVPR2016}, which is not mentioned in the DAVIS Benchmark, provided by the authors with the recommended parameter settings. 

Figure \ref{img:visual_comparison_vos} shows some examples of the obtained results. The quantitative comparison of these methods is shown in Table \ref{tab:vos}, indicating that our proposed method STCRF* exhibits the best performance on all the metrics at all the statistics. 
STCRF* achieves 0.714 for $\mathcal{J}(\rm{Mean})$, 0.674 for $\mathcal{F}(\rm{Mean})$, and 0.694 for $\mathcal{O}(\rm{Mean})$, while the best VOS methods achieve 0.558 (for FST~\cite{Papazoglou-ICCV2013}), 0.523 (for NLC~\cite{Faktor-BMVC2014}), and 0.537 (for NLC~\cite{Faktor-BMVC2014}), respectively.
STCRF* outperforms the compared VOS methods by a large margin on all the metrics. 
We can thus conclude that our proposed SOD method works even for VOS.
We note that DHS* is second best. 

\section{Conclusion} \label{section:conclusion}

Different from the still image, the video has temporal information and how to incorporate temporal information as effectively as possible is the essential issue for dealing with the video. This paper focused on detecting salient objects from a video and proposed a framework using STD features together with STCRF. Our method takes into account temporal information in a video as much as possible in different ways, namely, feature extraction and saliency computation. Our proposed STD feature utilizes local and global contexts in both spatial and temporal domains. The proposed STCRF is capable to capture temporal consistency of regions over frames and spatial relationship between regions. 

Our experiments show that the proposed method significantly outperforms state-of-the-art methods on publicly available datasets. We also applied our method to the video object segmentation task, showing that our method outperforms existing unsupervised VOS methods on the DAVIS dataset. 

Visual saliency is also used for estimating human gaze~\cite{Itti-PAMI1998}\cite{Harel-NIPS2006}\cite{Hou-CVPR2007}. For salient object detection, object boundaries should be kept as accurately as possible while for human gaze estimation, they are not.  Rather, gaze fixation point should be precisely identified and the area nearby the fixation point had better be blurred to present saliency using a Gaussian kernel, for example.
Applying our method directly to gaze estimation is thus not suitable.  However, the idea of combining local and global features
will be interesting even to gaze estimation.  Adapting our proposed method to gaze estimation in videos is left for future work.

\section*{Acknowledgment}

The authors are thankful to Gene Cheung for his valuable comments to improve the presentation of this paper.
This work is in part supported by JST CREST (Grant No. JPMJCR14D1) and by Grant-in-Aid for Scientific Research (Grant No. 16H02851) of the Ministry of Education, Culture, Sports, Science, and Technology of Japan.

\ifCLASSOPTIONcaptionsoff
  \newpage
\fi


\bibliography{short_bibtex}
\bibliographystyle{IEEEtran}

\end{document}